\documentclass[journal,onecolumn]{IEEEtran}

\usepackage{textcomp}

\usepackage{comment}

\usepackage{multirow}

\usepackage[noadjust]{cite}
\usepackage{xcolor}

\usepackage[T1]{fontenc} 
\usepackage{amsmath}
\usepackage[cmintegrals]{newtxmath}
\usepackage{bm}

\usepackage{algorithm}
\usepackage{algpseudocode}

\usepackage{array}

\usepackage{url}
\usepackage[hidelinks]{hyperref}
\usepackage[utf8]{inputenc}
\usepackage{graphicx}
\usepackage{amsmath}
\usepackage{booktabs}
\usepackage{subfigure}

\providecommand{\hypersetup}[1]{\relax}

\hypersetup{pdftitle={Bare Demo of IEEE\_lsens.cls for IEEE Sensors Letters},
pdfsubject={Typesetting},
pdfauthor={Michael D. Shell},
pdfkeywords={Class, IEEE, IEEE\_lsens, IEEE Sensors Letters, LaTeX, Typesetting, TeX}}

\usepackage{style}
\usepackage{bbm}
\begin{document}
\title{Towards Explainable Motion Prediction using Heterogeneous Graph Representations}

\author{\IEEEauthorblockN{Sandra~Carrasco~Limeros\IEEEauthorrefmark{1}\IEEEauthorrefmark{2}\IEEEauthorrefmark{3},
Sylwia~Majchrowska\IEEEauthorrefmark{2}\IEEEauthorrefmark{3},
Joakim~Johnander\IEEEauthorrefmark{3}\IEEEauthorrefmark{4}, 
Christoffer~Petersson\IEEEauthorrefmark{3}\IEEEauthorrefmark{5}, and
David~Fernández~Llorca\IEEEauthorrefmark{1}\IEEEauthorrefmark{6}
}\\
\IEEEauthorblockA{\IEEEauthorrefmark{1}Computer Engineering Department, Polytechnic School, University
of Alcala, Madrid, Spain}\\
\IEEEauthorblockA{\IEEEauthorrefmark{2}AI Sweden, Göteborg, Sweden}\\
\IEEEauthorblockA{\IEEEauthorrefmark{3}Zenseact AB, Göteborg, Sweden}\\
\IEEEauthorblockA{\IEEEauthorrefmark{4}Department of Electrical Engineering, Linköping University, Linköping, Sweden}\\
\IEEEauthorblockA{\IEEEauthorrefmark{5}Chalmers University of Technology, Göteborg, Sweden} \\
\IEEEauthorblockA{\IEEEauthorrefmark{6}European Commission, Joint Research Centre, Seville, Spain}
\\
sandra.carrascol@uah.es, sylwia.majchrowska@ai.se, joakim.johnander@zenseact.com, christoffer.petersson@zenseact.com, david.fernandez-llorca@ec.europa.eu

\thanks{Manuscript received November DD, 2022; revised Month DD, 2023.}
}

\maketitle

\begin{abstract}
Motion prediction systems aim to capture the future behavior of traffic scenarios enabling autonomous vehicles to perform safe and efficient planning. The evolution of these scenarios is highly uncertain and depends on the interactions of agents with static and dynamic objects in the scene. GNN-based approaches have recently gained attention as they are well suited to naturally model these interactions. However, one of the main challenges that remains unexplored is how to address the complexity and opacity of these models in order to deal with the transparency requirements for autonomous driving systems, which includes aspects such as interpretability and explainability. In this work, we aim to improve the explainability of motion prediction systems by using different approaches. First, we propose a new Explainable Heterogeneous Graph-based Policy (XHGP) model based on an heterograph representation of the traffic scene and lane-graph traversals, which learns interaction behaviors using object-level and type-level attention. This learned attention provides information about the most important agents and interactions in the scene. Second, we explore this same idea with the explanations provided by GNNExplainer. Third, we apply counterfactual reasoning to provide explanations of selected individual scenarios by exploring the sensitivity of the trained model to changes made to the input data, i.e., masking some elements of the scene, modifying trajectories, and adding or removing dynamic agents. The explainability analysis provided in this paper is a first step towards more transparent and reliable motion prediction systems, important from the perspective of the user, developers and regulatory agencies. The code to reproduce this work is publicly available at \url{https://github.com/sancarlim/Explainable-MP/tree/v1.1}.
\end{abstract}


\IEEEpeerreviewmaketitle

\section{Introduction}

\IEEEPARstart{A}{utonomous} vehicles (AVs) have to perform trajectory planning based on the global route and the local context. Trajectory planning can be applied in a safer and more efficient way if the system is able to anticipate future motions of surrounding agents \cite{Bahari2021}, as humans inherently do.

Motion prediction has recently gained significant attention within the research community since it is one of the key unsolved challenges in reaching full self-driving autonomy \cite{Ghorai2022}. The main goal of motion prediction is to determine a set of coordinates at a future point in time for an agent in the scene. 

\begin{figure}[t]
    \centering
    {\includegraphics[width=\linewidth]{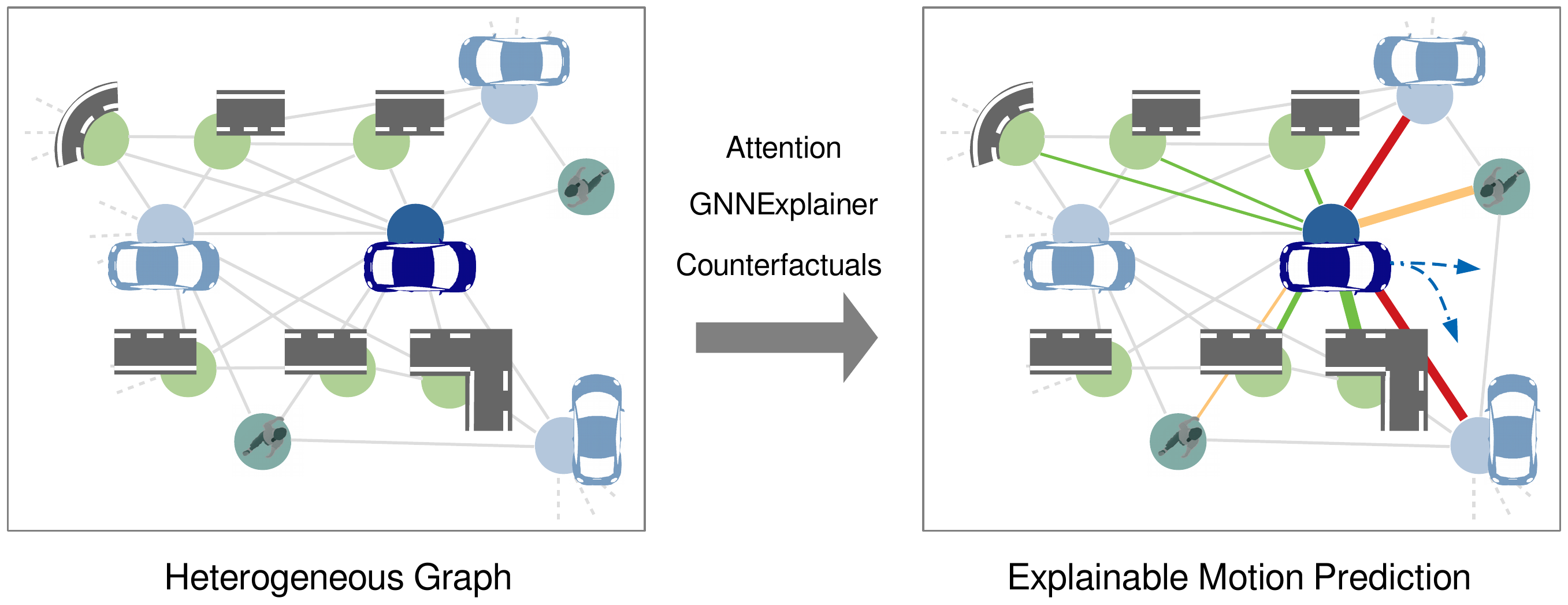}}
    \caption{Explainable Heterogeneous Graph-based Policy (XHGP). Explainable motion prediction is approached from three points of view: by visualizing the attention learned by the model, by using the GNNExplainer method, and by exploring counterfactual reasoning.   }
    \label{fig:abstract}
\end{figure} 

Among the different approaches, graphs are gaining attention since traffic scenarios can be naturally represented as a graph.  For example, by defining the different elements of the scene -- such as vehicles, pedestrians, or lane segments -- as nodes and their interactions as edges \cite{bib:Carrasco2021, deo2021multimodal, Gu2021DenseTNTET}.
Recently, there have been many studies on extending deep learning approaches for graph data, such as Graph Convolutional Networks \cite{GNN} and Graph Attention Networks \cite{GAT}. They use neural networks and attention mechanisms to incorporate the feature information  of nodes in a graph -- such as position or velocity -- as well as the structure information -- such as the distance or the type of relation -- and pass these messages through the edges of the graph. In this work, we focus on these types of approaches to deal with complex and highly interactive scenarios with different road topologies and multiple agents.  

How to specify the prediction itself is a key issue that affects interpretability.  Multi-modal motion prediction tries to capture the underlying  distribution of future motions handling multiple predictions with an associated probability. This is a much more sensible approach than deterministic prediction models, since in complex real-world scenarios, there is a high uncertainty of traffic behavior and many potential situations. Moreover, multi-modal approaches have shown to produce more balanced and robust performance, and to improve the interpretability of outputs, especially when accompanied by an intent detection system and appropriate visualization of the predictions \cite{Carrasco2022}. 

Yet the complexity and opacity of the approaches used to deal with motion prediction is a problem that impacts on users, developers and approval authorities, including auditors and investigators. Regulators are increasingly aware of these intrinsic limitations of Artificial Intelligence (AI), and the policy trend is to impose transparency, accountability and human oversight obligations on the providers of these systems \cite{AIAct}, or even explainability requirements \cite{US-AAA2022}, in line with the ethical design requirements of trustworthy artificial intelligence \cite{HLEGAI}.

The transparency requirements for autonomous driving establish the need for humans to understand and trace the decisions made by an autonomous vehicle, including different aspects such as traceability, data logging and explainability \cite{Llorca2023}. In particular, explainability is beneficial for several stakeholders. First, it can help internal and external users to understand the autonomous vehicle's operation, allowing for trustworthy human-vehicle interaction. It also helps to improve testing, certification and auditing procedures \cite{Izquierdo2022}. Finally, developers themselves can benefit considerably from explainable systems \cite{Llorca2021}.

Since an autonomous vehicle is a very complex system, incorporating multiple AI systems to address different problems (e.g. motion prediction), it can be assumed that the explainability of each of these systems is a pre-condition for explainability at the level of the autonomous vehicle. In this particular case, explaining the predictions of deep graph models includes determining which input edges -- interactions --, input nodes -- agents --, or node patterns are the most important. In addition, explainability methods aim to find the graph patterns that maximize the prediction of a certain agent or that is the most relevant for a certain scenario.

To the best of our knowledge, explainability in graphs applied to traffic scenarios has not yet been explored. This work is a first attempt to explore and adapt different explainability approaches to provide useful explanations of the predictions established by the graph-based system. As a step towards explainable multi-modal motion prediction in highly complex and interactive scenarios, our contributions are three-fold:

\begin{itemize}
    \item  We propose a new method based on  graph-traversals~\cite{deo2021multimodal}, in which road and dynamic agents are jointly modeled in a heterogeneous graph, as a means to have a more explainable model.  We evaluate our model on the challenging large-scale NuScenes dataset.
    \item We provide an analysis of different explainability methods applied to GNNs in the context of multi-modal motion prediction. 
    \item We perform a qualitative and quantitative assessment of the provided explanations using diverse and challenging NuScenes dataset. 
\end{itemize}

\section{Related works}

Deep neural networks have been successful in a wide range of tasks, including motion prediction \cite{Huang2022}. However, a well-known drawback with deep neural networks is that it is difficult to explain their predictions. Without reasoning about the underlying mechanisms behind the predictions, deep models cannot be fully trusted, which precludes their use in critical applications due to fairness, privacy, and safety concerns.
The Defence Advanced Research Project Agency (DARPA) coined the term Explainable AI (XAI) as a new research direction to address this weakness \cite{gunning2016explainable}.

The terms of interpretability and explainability are usually interecheangeably used. However, although closely related, some works try to clarify the difference between these concepts, \cite{towards-rigorous, article-rudin, explaining}, but none of them provide a rigorous formal mathematical definition. \textit{Interpretability} is defined by \cite{towards-rigorous} as  “the ability to explain or to present in understandable terms to a human” meaning that the cause and effect can be determined. In this work, we follow \cite{article-rudin} and consider a model to be interpretable if the model itself can provide understandable interpretations of its predictions, such as a decision tree model. However, most machine learning models are designed with specific accuracy as a goal and do not have interpretability constraints. 

In contrast,  \textit{explainable} models are those that are still a black box and require post hoc techniques to explain its predictions. According to \cite{explain}, an explanation must provide at least one of the following:
the main factors in the decision process; whether the decision would have changed if the factors had been modified;
whether two similar  cases result in different decisions, or vice versa.

In this section, we first review recent advances in graph-based motion prediction. Next, we review explainability in GNNs. Finally, we identify previous attempts to deal with interpretable and explainable motion prediction. 

\subsection{GNN-based motion prediction}
\label{subsec:mm-pred}
Graph Neural Networks (GNNs) have been widely used to capture and model the underlying spatial and temporal relationships of the agents in a traffic scene. Some approaches propose the use of the GNN model to encode the spatial interactions between traffic agents and then attaching some recurrent \cite{GRIP2019, Zhou2021, Mo2021, Tang2022, Zhang2022-T-IV} or attention-based \cite{Zhang2022-T-ITS} system to model the temporal relationships and generate the predicted trajectories. However, the temporal information can be intrinsically captured by the graph model by adding self-connections to the nodes as temporal edges \cite{bib:Carrasco2021}.

Apart from the spatial and temporal information of the agents, prior knowledge of the road structure plays a fundamental role. Most self-driving cars have access to high definition (HD) vector maps, which
contain detailed geometric information, such as roads, lanes, intersections, crossings, traffic signs, and traffic lights. The simplistic way to encode road and motion information is rasterization~\cite{SpAGNN}. In such a representation, different semantics are encoded in separate channels, which facilitates the learning of the convolutional neural network (CNN). However, rasterization may lose useful information like lane topologies and has difficulty to capture long range interactions. A more efficient solution is to represent structured HD maps and agents using polylines. Gao et al.~\cite{vectornet2020} treat a map as a collection of polylines and use self attention-subgraphs to encode them.
Discretization of the scene context in form of vectors was adopted in subsequent works~\cite{tnt,Gu2021DenseTNTET,Zhao2022} because it provides a uniform representation with a lower computational cost. In addition, it allows interpretable analysis of the model's behavior and enables counterfactual predictions that condition hypothetical "what-if" polylines~\cite{Khandelwal2020WhatIfMP}.

Other ideas involve building additional structures. For example, LaneGCN~\cite{Lanegcn} builds an extra graph of lanes and conducts convolutions over the graph. LaneRCNN~\cite{LaneRCNN} encodes both actors and maps in an unified graph representation, which is even more structured. Both methods encode the input context into a single context vector. It is then used by the multi-modal prediction header to derive multiple likely future trajectories. The predictive header therefore has to learn a complex mapping, which often leads to predictions that get out of the way or violate traffic rules.

Recently introduced Prediction via Graph-based Policy (PGP) model~\cite{deo2021multimodal} achieves great scene compliance due to the use of a lane-graph traversals approach. It selectively aggregates scene context based on path traversals sampled from a learned behavior cloning policy, capturing the lateral variability of the output distribution. To model longitudinal variability, a sampled latent variable was added, enabling prediction of different trajectories for a same traversal.

\subsection{Explainability with Graph Neural Networks } 

Explainability in GNNs can be taxonomized in instance-level and model-level methods. The former provide input-dependent explanations, while the latter explain the general behavior of a GNN at a higher level.  XGNN \cite{Yuan_2020} proposes to explain GNNs by learning to generate graphs that achieve optimal prediction scores according to the GNN model to be explained.
Instance-level methods have been studied much more extensively and can be divided into four categories:
\begin{itemize}
    \item \textbf{Gradient-based methods} use gradient or hidden features as proxies for input importance to explain the predictions through back-propagation. \cite{8954227, SA}
    \item \textbf{Perturbation-based methods} study the change of output variations for different input perturbations. \cite{GNNEXp, PGExplanier, GraphMask, casual, CFGNN}
    \item \textbf{Decomposition-based methods} track the contribution of individual graph components to the final prediction by decomposing the original model predictions into different terms that are considered as importance scores of the input features. \cite{Schnake_2021,decomposition,8954227}
    \item  \textbf{Surrogate methods} use a simpler and more interpretable surrogate model to approximate the underlying GNN  predictions as accurately as possible. \cite{PGMExp, GraphLime}
\end{itemize}

\subsection{Interpretable and explainable motion prediction}

The need to understand the importance of road users' interactions was previously highlighted by the authors of the Socially-COnsistent and UndersTandable (SCOUT) Graph Attention Network (GAT)~\cite{bib:Carrasco2021}. This was investigated using the Integrated Gradients~\cite{ig} technique and visualization of learned attention. The illustrations provided for four carefully selected scenarios showed the importance of spatial representation of the nearest traffic participants. The use of the learned attention as a mechanism to generate explainable motion predictions was also proposed for transformer-based models \cite{Zhang2022-TRC}. 

The explainability of motion prediction was recently explored by use of conditional forecasting~\cite{Khandelwal2020WhatIfMP}. The authors of \textit{What if Motion Prediction?} (WIMP) model~\cite{Khandelwal2020WhatIfMP} developed an iterative, graphical attention approach with interpretable geometric (actor-lane) and social (actor-actor) relationships that support the injection of counterfactual geometric targets and social contexts. In this way, the model can make different predictions based on injected/removed lanes and actors. The proposed method supports the study of hypothetical or unlikely scenarios, so-called counterfactuals. This capability can be further used in the planning process to include only relevant features in the calculations or to examine and evaluate the model's predictions.

In~\cite{Zhang2022}, the authors investigated the robustness of motion prediction for AV by proposing a new adversarial attack that disrupts normal vehicle trajectories to maximize prediction error. This study focuses on 3 predictive models and 3 trajectory datasets. The evaluation showed that the prediction models are generally susceptible to perturbation by adversaries and can result in unsafe AV behavior, such as harsh braking. The authors highlighted the need to evaluate the scenario of traffic prediction models. Data augmentation and trajectory smoothing were proposed as mitigation methods. This reduced the prediction error under attacks by 28\%.

To our knowledge, explainability in graph-based motion prediction for autonomous vehicles has not yet been explored. This work is a first attempt to move towards explainable graph-based motion prediction for autonomous driving. 

\section{Methodology}

We propose a novel approach for motion prediction named Explainable Heterogeneous Graph-based Policy (XHGP). This model extends the lane-graph traversals idea presented in~\cite{deo2021multimodal} to jointly model lanes and different types of dynamic agents in a heterogeneous graph~\cite{iehgcn}. 

\subsection{Problem formulation} 
\label{subsec:definition}

The goal is to estimate future trajectories of vehicles of interest from their past trajectories and scene context. Input features describe the past trajectory of the vehicle of interest as well as the scene context.

\begin{figure}[!t]
\centering
    \subfigure[]
    {
     \label{subfig:pf1}
    \includegraphics[width=.5\linewidth]{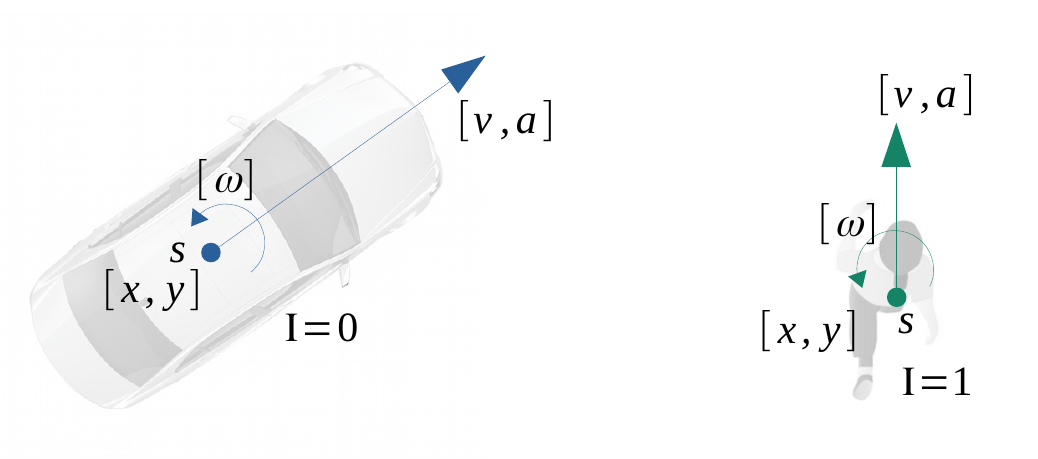}
    }\hspace{10mm}
    \subfigure[] 
    {
        \label{subfig:pf2}
        \includegraphics[width=.3\linewidth]{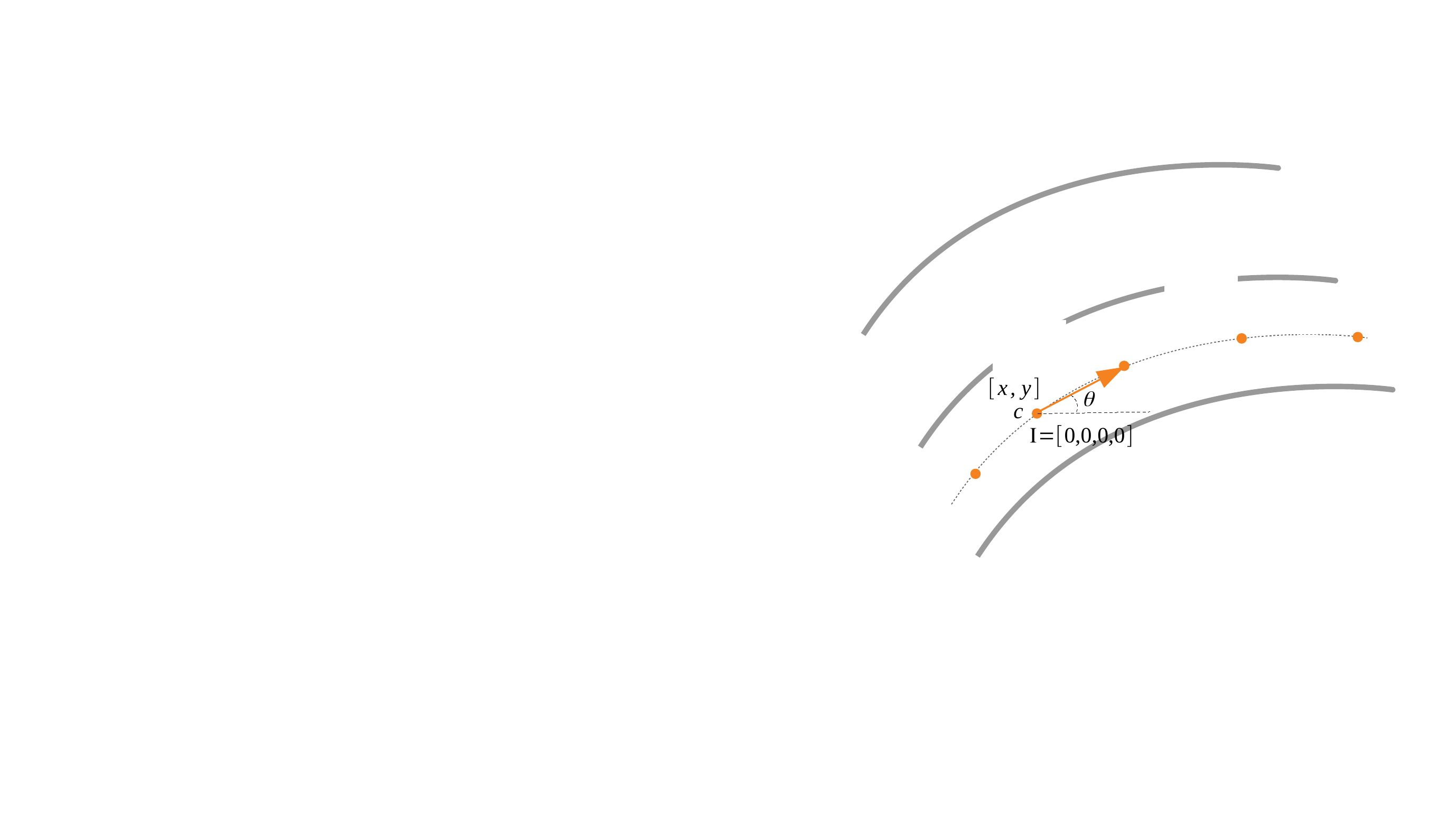}
    }
\caption{Illustration of the state variables (a) of an agent and (b) of a lane centerline point.}
\label{fig:pf}
\end{figure}

One the one hand, let $s_i^t\in\mathbb{R}^6$ be the state of agent $i$ at time $t$, including its two-dimensional position in the BEV-plane, $\tau_i^t=[x_i,y_i]$, as well as the velocity $v_i$, acceleration $a_i$, yaw rate $\omega_i$, and a flag $I_i$ indicating the type of agent -- pedestrian ($I_i=1$) or vehicle ($I_i=0$). Therefore $s^t_i=[x_i,y_i,v_i,a_i,\omega_i,I_i]$ (see Fig. \ref{subfig:pf1}).
On the other hand, lane centerlines are represented by a segment of fixed length, each segment consisting of a sequence of $N$ points. Each lane point $n$ is defined by a vector $c_n\in\mathbb{R}^7$ which includes its two-dimensional position in the BEV-plane $[x_n, y_n]$, the yaw $\theta_n$, and a semantic 4D binary vector $I_n$ indicating whether the point belongs to a stop line, turn stop, crosswalk, or traffic light. That is, $c_n=[x_n, y_n, \theta_n, I_n]$ (see Fig. \ref{subfig:pf2}).

The aim is to predict $\tau_i^t$ for future time steps $T_\text{obs}<t\le T_\text{pred}$, being  $ T_\text{pred}=6$ at 2Hz.

\subsection{Multi-modal motion prediction model}
Future behaviors of road agents are inherently uncertain and conditioned on the scene context, including road topology and social interactions with other road agents. According to this idea, we explored different model architectures to explicitly capture these two interactions with respect to the agent of interest. 

We extend \cite{bib:Carrasco2021},  an attention-based GNN that uses a flexible and generic representation of the scene as a graph for modelling interactions predicting socially-consistent trajectories by adding the scene context as a rasterized HD Map and a multimodal prediction header.  However, rasterization is inefficient and prone to information loss. Recently, lane-graph based methods have shown better performance and efficiency.
On the other hand, our experiments with multi-modal autoencoders based on mixture density networks and on variational autoencoders showed to be prone to mode collapse, with different modes being different in speed profile, landing on a single path. The approach taken in \cite{deo2021multimodal}, based on lane-graph traversals conditioning, has shown promising results. They leverage the strong inductive bias that lanes provides to the network, which capture both the direction of traffic flow and the legal routes for each agent in the scene.

\begin{figure*}[t]
    \centering
    {\includegraphics[width=\linewidth]{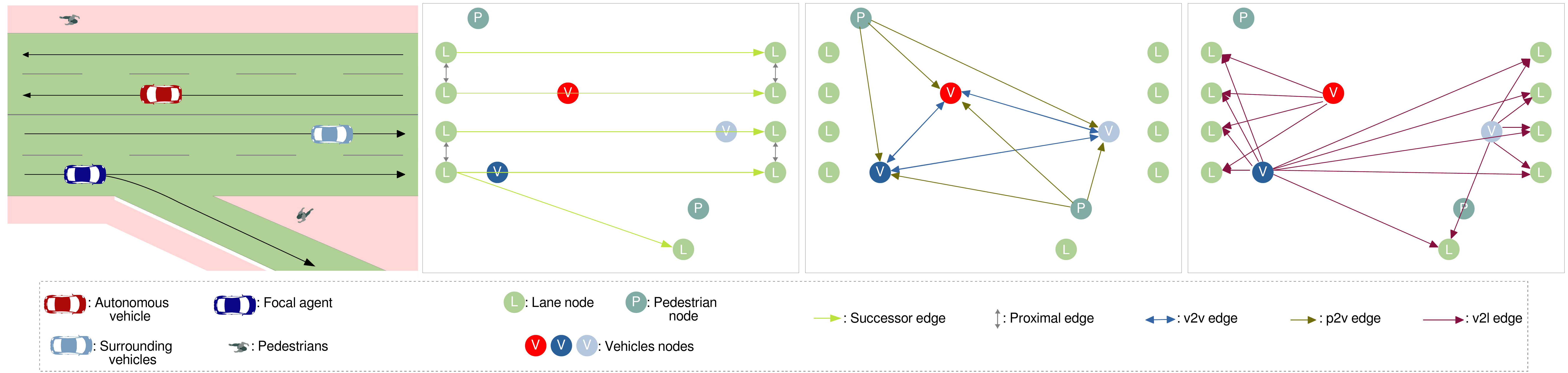}}
    \caption{Example of building a heterogenous graph from a traffic scene. Nodes include vehicles, pedestrians and lanes. Directed and undirected edges include multiple types of interactions.}
    \label{fig:graph1}
\end{figure*} 

In this work, we model traffic scenarios as graphs, with nodes representing the different agents and static objects in the scene, and edges representing their interactions. This is a more natural and flexible way of encoding the scene context than rasterization, since it can handle different input sizes and achieve invariance to input ordering. 
Using a generic representation of the scene as a graph allows the model to be flexible to different environments and road topologies and to harness the expressive power of GNNs. Additionally, graph-based representations contain important topology information, which makes them more interpretable than other DL approaches in this scenario, being able to directly retrieve the importance of each interaction in the graph.

\textbf{Graph definition.}
We propose to model the whole scene as an heterogeneous graph including both scene and agent context information (see Fig. \ref{fig:graph1}). As stated in \cite{Zhang2022-TRC}, the use of heterogeneous graphs makes it possible to merge information on road topology and traffic rules, as well as to rationalize the implicit interactive information of the agents. Hence, we define the whole scene context as a directed heterogeneous graph $\mathcal{G}=\{\mathcal{V},\mathcal{E},\mathcal{T},\mathcal{R}\}$. Each node $v \in \mathcal{V}$  has a \emph{type}, formally defined by a mapping function $\tau(v):\mathcal{V}\rightarrow\mathcal{T}$, and each edge  $e \in \mathcal{E}$  has a \emph{type}, formally defined by a mapping function $\phi(e): \mathcal{E} \rightarrow \mathcal{R}$. Therefore, $\mathcal{T}$ denotes the set of node types and $\mathcal{R}$ the set of relations types. 
$\mathcal{V}^\Omega$ is the set of nodes of type $\Omega \in \mathcal{T}$ and $\mathcal{N}^\Omega = \{ \Gamma | \Gamma, \Omega \in \mathcal{T}, \langle \Gamma, \Omega \rangle \in \mathcal{R} \}$ denotes the neighborhood of $\Omega$, being $\Gamma \in \mathcal{N}^\Omega$ a neighbor type of $\Omega$. The interaction from $\Gamma$ to $\Omega$ is denoted as $\langle \Gamma, \Omega \rangle$. 

In this work, $\mathcal{T}=\{\text{lane}, \text{vehicle}, \text{pedestrian}\}$, and  $\mathcal{R}=\{\text{succ}, \text{prox}, \text{v2l},\text{v2v}, \text{p2v}\}$ -- defining successor, proximal, vehicle-to-lane, vehicle-to-vehicle and pedestrian-to-vehicle interactions, respectively. 
Following \cite{deo2021multimodal}, each lane node, $v\in\mathcal{V}^\text{lane}$, represents a lane centerline segment of 20m length. Each node is defined by  a sequence of points with its feature vectors $c^l = [c_1^l,\dots,c_N^l]$ as described in \ref{subsec:definition}.
Successor edges connect lane nodes along the lane. Proximal edges connect neighboring lane nodes that are within a distance threshold of each other and their yaw angle difference is within a threshold.
We consider all agents and lanes within a specific area around the focal agent. Concretely, 50m laterally, 20m behind and 80m in front of the agent.
For $E^{v2l}$, we consider all lane nodes within 10m around the particular vehicle. The focal agent is connected to all lanes within the predefined region.
For $E^{v2v}$ and $E^{p2v}$, we define an interaction when the euclidean distance between two agents is lower than 20m. 

We define each relation with the adjacency matrix $A^{\Omega-\Gamma} \in \textbf{R}^{|\mathcal{V}^\Omega| \times |\mathcal{V}^\Gamma|}$, where the element $s_{ij}$ at row $i$ and column $j$,
represents whether two nodes are connected. Note that the adjacency matrix may not be square for heterogeneous relations, since $|\mathcal{V}^\Omega|$ may not be equal to $|\mathcal{V}^\Gamma|$.

\textbf{Scene and agent context encoding.}
Each object type is first encoded using a gated recurrent unit (GRU). We use a separate encoder for the focal agent, surrounding agents -- where the type of agent (vehicle or pedestrian) is indicated in the input features --, and lanes nodes. Then, we obtain the encoding $Z^\text{v}$,  $Z^\text{p}$, and $Z^\text{l}$.

After encoding and projecting all the hidden representations of neighbor objects into a common semantic space, we employ an attention-based \textbf{\textit{object-level}} aggregation similar to that in~\cite{bib:Carrasco2021, GAT}, instead of the statistically \textit{row-normalized} convolution operation. For each canonical edge type in $\mathcal{E}$, a subgraph is considered for the aggregation, with source  $\mathcal{V}^\Omega$, destination $\mathcal{V}^\Gamma$ and relation $\mathcal{R}^{\Omega-\Gamma}$. We then compute a pair-wise un-normalized attention score between each two neighbors in the form of additive attention, and normalize them across all the neighborhood of $\Omega$ using the softmax function, 
    \begin{equation}\label{eq:object-att}
    \alpha_{ij}  = \text{Softmax}_j ( a^T \cdot \text{LeakyReLU} (  [W  z_i  ||W  z_j ]  ) )\enspace,
     \end{equation}
where $z$ is the node encoding, $W$ is the weight matrix that parametrizes a shared linear transformation applied to every node, $a$ defines the self-attention mechanism, $\cdot^{T}$ represents transposition, and || is the concatenation operation.

\begin{figure*}[t]
    \centering
    {\includegraphics[width=0.9\linewidth]{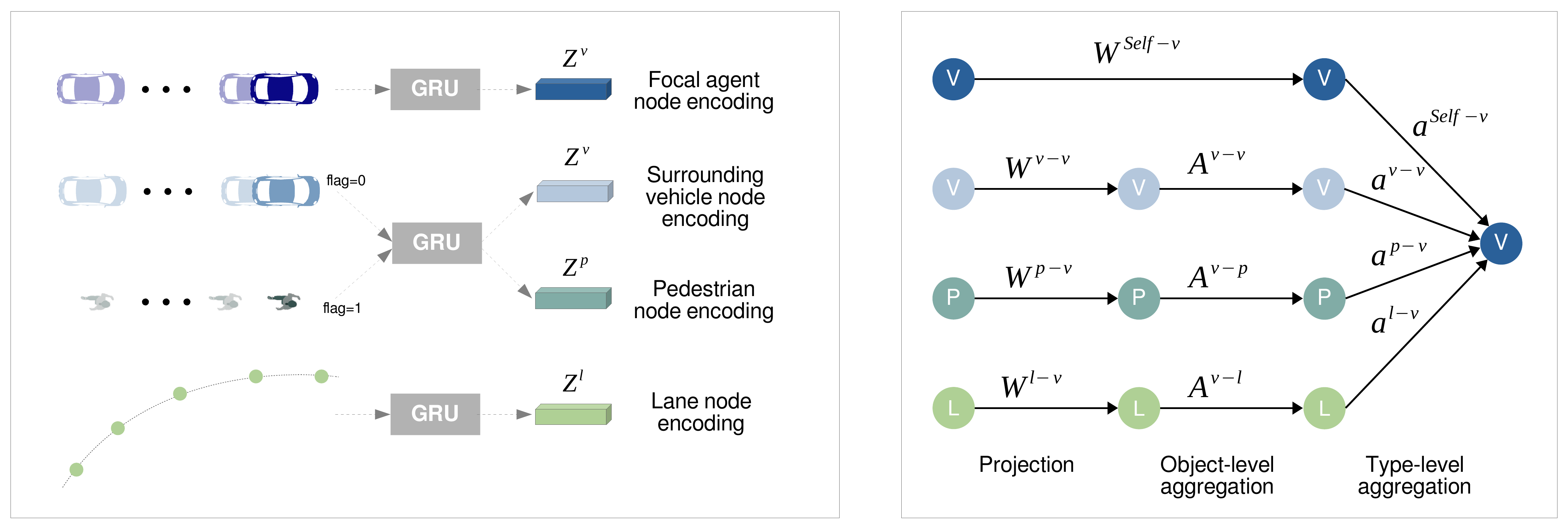}}
    \caption{Left: lane and agent node encoding using different Gate Recurrent Units for the focal agent, surrounding vehicles, and lane nodes. Right: object- and type- level aggregation. $W$ projects the representation of each node to a common semantic space. $A$ is the adjacency matrix used for object-level aggregation. $a$ is the attention mechanism for type-level aggregation.}
    \label{fig:graph2}
\end{figure*} 

The attention score indicates the importance of node $j$ features to node $i$, which allows each node to attend every other node in its neighborhood. This score is later visualized to understand which interactions are most important.
The aggregation of the neighbors embeddings is similar to GCN~\cite{DBLP:conf/iclr/KipfW17}, where all the neighbors in the subgraph are aggregated together scaled by the attention score. 

To learn more powerful representations for each node type, \cite{iehgcn} proposes \textbf{\textit{type-level}} attention to fuse representations from different types of neighbor objects. The intuition is that for each type of object, the information coming for different types of neighbor objects could not have the same importance. For example, for objects of type \textit{lane}, the relation \textit{vehicle-to-lane} could be more relevant for the final prediction than the relation \textit{proximal}. For objects of type \textit{vehicle}, the relation \textit{vehicle-to-vehicle} in general should be more relevant than \textit{pedestrian-to-vehicle}.
Hence, before aggregating the representations from the object-level attention aggregation, we first learn the importance for the different types of neighbors using scaled dot product attention~\cite{NIPS2017_3f5ee243}. 
In order to compute the importance of the representations of neighbor types, including the self representation, for $\mathcal{V}^\Gamma$, we linearly project the hidden representation from the previous step,  $\{\textbf{H}^{\Omega-\Gamma}\} \cup \{\textbf{H}^{\Gamma}\}$, into keys, and  $\{\textbf{H}^{\Gamma}\}$ into the query.  
The values will be the output from the previous step. 

Then, the type-level attention is computed as follows:

\begin{equation} \label{eq:type-att1}
    e^{\Gamma} = \text{ELU} ( [\textbf{H}^{\Gamma} \cdot \textbf{W}_k^\Gamma || \textbf{H}^{\Gamma} \cdot \textbf{W}_q^\Gamma ] \cdot  w^\Gamma)
\end{equation}
\begin{equation} \label{eq:type-att2}
      e^{\Omega-\Gamma} = \text{ELU} ( [\textbf{H}^{\Omega-\Gamma} \cdot \textbf{W}_k^\Gamma || \textbf{H}^{\Gamma} \cdot \textbf{W}_q^\Gamma ] \cdot w^\Gamma),\quad \Omega \in \mathcal{N}^\Gamma
\end{equation}
Finally, a softmax function is applied to get the normalized attention coefficients, $a_i^{\Gamma}$ and $a_i^{\Omega-\Gamma}$. These attention scores are then used to compute the higher level representation of  $\mathcal{V}^{\Gamma}$. For each element $i$ in  $\mathcal{V}^{\Gamma}$,

\begin{equation} \label{eq:aggregation}
    \hat{\textbf{H}}^{\Gamma}_i = \text{ELU} ( a_i^{\Gamma} \cdot \textbf{H}_i^\Gamma + \sum_{\Omega \in \mathcal{N}^\Gamma} a_i^{\Omega-\Gamma} \cdot \textbf{H}_i^{\Omega-\Gamma})
\end{equation}

We apply two layers of the defined GNN to aggregate the two-hop neighborhood information. The node encoding size and the attention size is 32.

\textbf{Graph traversal and decoding.}
 The final representation for the focal agent, $\hat{\textbf{H}}^v_0$,  is concatenated with its encoding previous the aggregation, $Z^\text{v}_0$.  This output, together with the representation for the lane nodes,  $\hat{\textbf{H}}^l_0$, is used to learn a policy for graph traversal using behavior cloning. The final decoding is conditioned on path traversals, following~\cite{deo2021multimodal}.  

\section{Towards explainable motion prediction}

\subsection{Attention visualization}

As explained in the previous section, the  scene context encoding is performed using two types of attention: object-level and type-level attention. The former reflects the importance of each neighboring node in the aggregation, i.e., how each interaction affects the prediction. The latter represents the importance of each type of interaction. We aggregate both attention scores and visualize them to understand which interactions in the scene are most relevant for the final prediction.

\subsection{Post-hoc explainability methods: GNNExplainer}

Existing methods adapting or specifically designing the explanation methods for GNNs have shown promising explanations on multiple types of graph-structured data~\cite{Jaume2020TowardsEG,Rao2021QuantitativeEO,patology,Zhdanov2022InvestigatingBC}. 
In particular, perturbation methods involve learning or optimization~\cite{GNNEXp, PGExplanier, GraphMask, casual, CFGNN} and, while bearing higher computational costs, generally achieve state-of-the-art performance in terms of explanation quality. 
These methods train local post-hoc interpretable models on top of the explained predictive model.
One of the most interesting approaches is GNNExplainer~\cite{GNNEXp} which requires training or optimizing an individual explainer for each data instance, i.e., a graph or a node to be explained. 

GNNExplainer is a graph pruning based explainability technique, which can be used with any type of GNN. The explainer tries to find the minimum sub-graph $G_s \in G$ such that the model prediction is retained. The task is formulated as an optimization problem that learns a mask to activate or deactivate parts of the graph. The initial formulation~\cite{GNNEXp} was developed for explaining node classification tasks and is based on edge masking. However, other works~\cite{Jaume2020TowardsEG} utilised the technique also to learn a mask over the nodes instead of edges. 

Formally, GNNExplainer aims to maximize mutual information between a compact sub-graph $G_s$ and a base graph $G$ by learning an edge mask and a feature mask in an optimisation process~\cite{GNNEXp}. The undoubted advantage of the methodology is its versatility, that is, its adaptability to problems other than graph or node classification or link prediction, for which it has been used in previous studies. We utilised the approach to work on heterogeneous graphs and learn the importance mask of its edges (traffic agents interactions) for the prediction of the focal agent trajectory for each timestamp in the given scenario. We intend the explanations to be as compact as possible, while providing the same prediction as the original graph.
Heuristically, we enforce these constraints by optimizing the following objective function
\begin{equation}
\begin{split}
    \mathcal{L} = \mathcal{L}_{dist}(y,\hat{y}) &+ \alpha_1||M||_1 + \alpha_2\mathcal{H}(M) \\
                &+ \beta_1||F||_1 + \beta_2\mathcal{H}(F)),
\end{split}
\end{equation}
where $\mathcal{L}$ is the loss function measuring distance between original and predicted trajectory of interest (usually the most probable first mode), $\hat{y}$ is the original model prediction, $y$ is the model prediction with the edge ($M$) and feature ($F$) mask applied, $\mathcal{H}$ is the entropy function. To measure the distance between different predicted trajectories we use the same metric as for the predictive model.
In the end we obtained the edge weights, which we were able to compare with the attention learned by the GNN encoder of the XHGP model.

\subsection{Counterfactuals}

Mentioned above methods are not counterfactual by their nature, meaning that they do not explain how to achieve an alternative outcome from the predictive model. 
Counterfactuals about hypothetical occurrences are increasingly used in different explainable AI applications. A counterfactual explanation describes a cause and effect situation in the form \textit{If X had not occurred, Y would not have occurred}. The name \textit{counterfactual} comes from imagining a hypothetical reality that contradicts the reality. Counterfactual reasoning can be used to explain the predictions of individual instances made by a model by modifying the cause (input features) of that prediction. This makes counterfactuals  one of the most intuitive explanation for humans since, first, reasoning about cause and effect in a situation comes very natural to humans, and, second, they focus on a specific instance and a small input change. In our study we presented how changing different part of the modelled traffic scenario can affect the final prediction.

Our method models the entire scene as a heterogeneous graph, including interactions between the road network and dynamic agents. This architecture is more explainable in its nature than most DL-based approaches, since it allow us to capture interpretable geometric and social relationships. Besides, it makes it easier to efficiently apply counterfactual reasoning.  The surrounding context of the focal agent can be manipulated to ablate specific social or road influences, as well as to condition upon unobserved hypothetical agents. 

These counterfactual explanations can be used not only to explain, but also to evaluate predictive models that rely on their scene context 
and as a measure of robustness and generalizability. We want our model to produce sensible predictions when conditioned on unlikely extreme inputs, demonstrating that the model has learned a powerful causal representation of driving behavior. 
 This capability is not only useful for explainability, but could also be exploited by a subsequent decision making system or subsequent planner to reason about social influences from occluded regions.

\section{Evaluation framework} 

\subsection{Dataset}

In our experiments, the reference dataset is the public, large-scale nuScenes dataset~\cite{nuscenes2019} for autonomous driving developed by the Motional team.
The 1000 scenes of 20 seconds each, with ground-truth annotations and HD maps, were collected in Boston and Singapore, where right-hand and left-hand traffic rules apply respectively. 
In our prediction scenario, we rely on 2 seconds of dynamic agents' history and the map features to predict the next 6 seconds. 
Given the geographic diversity of the data and comprehensive representation of complex scenarios such as turns and intersections, nuScenes presents one of the most challenging prediction benchmarks.
For presented results, the challenge split from the nuScenes motion prediction benchmark~\cite{nuscenes2019} was used.

\subsection{Evaluation metrics }
\label{sec:metrics}

\subsubsection{Quantitative metrics for motion prediction}
\label{sec:model_metrics}
first, we perform a quantitative evaluation of the motion prediction model using nuScenes motion prediction benchmark. 
We output 10 predictions for the focal agent, 6s into the future at 2Hz, along with the probability that the agent follows that trajectory. The metrics used in the nuScenes benchmark to measure the degree to which this proposed set of trajectories matches the ground-truth are as follows:
\begin{itemize}
    \item Best-of-K Average Displacement Error (minADE): The minimum point-wise L2 distance to the ground-truth trajectory over all predicted trajectories.  
    \item Best-of-K Miss Rate (MR): percentage of predictions whose maximum pointwise L2 distance between the prediction and ground-truth is greater than 2.0m. 
    \item Off-road rate:  computes the fraction of trajectories that are off-road, outside the drivable area. 
\end{itemize}

\subsubsection{Quantitative metrics for graph explainability}
\label{sec:gnnexp_metrics}
in order to evaluate the GNN explanation method, it is necessary to qualitatively examine the results for each input example. 
This would be extremely time-consuming with only visualizations.
In addition, this type of evaluation is highly dependent on people's subjective understanding, resulting in a biased and unfair evaluation. Evaluation metrics are a good alternative to study these explanation methods. In this section, we introduce several recently proposed evaluation metrics~\cite{Yuan2022ExplainabilityIG}.

\textbf{Sparsity.} 
Good explanations should capture the most important input features and ignore the irrelevant ones.
The metric \textit{Sparsity} measures the fraction of features selected as important by the explanation method~\cite{8954227}.
Formally, for given $i-\mathrm{th}$ input of the graph $G_i$ and its importance weights $m_i$, the \textit{Sparsity} metric can be computed as
\begin{equation}
    \text{Sparsity} = \frac{1}{N}\sum_{i=1}^N{1-\frac{|m_i|}{|M_i|}},
\end{equation}
where $N$ is the number of graphs, $|m_i|$ denotes the number of important features (i.e., nodes, edges, node features) identified as $m_i$, and $|M_i|$ -- the total number of features in $G_i$. Higher values of the metric indicate that explanations are more sparse and tend to capture only the most important input information.

\textbf{Fidelity.} 
The explanations provided should be faithful to the model, meaning that they should identify input features that are important to the model, not (only) to the human. To evaluate this, the \textit{Fidelity+}~\cite{8954227} and \textit{Fidelity-}~\cite{Yuan2022ExplainabilityIG} scores were introduced. 
The \textit{Fidelity+} is defined as the difference of accuracy (or predicted probability) between the original
predictions and the new predictions after masking out important input features. For the given $i-\mathrm{th}$ input graph $G_i$, the score can be computed as
\begin{equation}
    \text{Fidelity+} = \frac{1}{N}\sum_{i=1}^N{\left(1-\mathbbm{1}(y_i^{1-m_i} = y_i)\right)},
\end{equation}
where $y_i$ is the original prediction of graph i. $1 - m_i$ means the complementary
mask that removes the important input features, and $y_i^{1-m_i}$ is the prediction when feeding the new graph into the trained GNN-based $f(\cdot)$. The indicator function $\mathbbm{1}(y_i^{1-m_i} = y_i)$ returns 1 if $y_i^{1-m_i}$ and $y_i$ are equal and returns 0 otherwise. For GNNExplainer~\cite{GNNEXp}, the importance scores are continuous values, and hence the importance map $m_i$ can be obtained by normalization and thresholding or ranking~\cite{Yuan2022ExplainabilityIG}. 
For \textit{Fidelity+} higher values indicate better explanation results and more discriminative features are identified.

In contrast, the metric \textit{Fidelity-} studies prediction change by keeping important input features and removing unimportant features, and is defined as
\begin{equation}
    \text{Fidelity-} = \frac{1}{N}\sum_{i=1}^N{\left(1-\mathbbm{1}(y_i^{m_i} = y_i)\right)},
\end{equation}
where $y_i^{m_i}$ is a new predicate based on the explanation of $m_i$ for $G_i^{m_i}$ being a new graph by preserving only the important features of $G_i$. For \textit{Fidelity-}, lower values indicate less importance information are removed so that the explanations results are better.

\section{Results}

\subsection{Quantitative evaluation of multi-modal motion prediction}
We compare our method against state-of-the-art models on \href{https://eval.ai/web/challenges/challenge-page/591/leaderboard/1659}{nuScenes benchmark}. Results are shown in Table~\ref{tab:mp}.
This method achieves similar results as the original PGP. The results described for PGP are obtained by reproducing their work, which improve on those presented in the original paper. In the table, we detailed both the results presented in their work and those obtained with the new implementation ($\cdot^{\dag}$). Our approach, however, presents a more explainable architecture with the ensuing benefits outlined above. 
Both these models achieve best results in nuScenes motion prediction leaderboard~\cite{nuscenes2019}.

\begin{table}[!ht]
\centering
\caption{Comparison to the state-of-the-art on nuScenes benchmark. We change the PGP encoding, modeling the entire scene as a heterograph, to make it compatible with GNN explanability methods without any loss of performance.}
\label{tab:mp}
\begin{tabular}{@{}c|ccccc@{}}
\toprule  
Method               & \multicolumn{2}{c}{MinADE}                         & \multicolumn{2}{c}{MissRate}                       & Off-road         \\
\multicolumn{1}{c|}{} & \multicolumn{1}{c}{K=5} & \multicolumn{1}{l}{K=10} & \multicolumn{1}{l}{K=5} & \multicolumn{1}{c}{K=10} & \multicolumn{1}{c}{} \\ \midrule

Trajectron++~\cite{traj++}   & 1.88      & 1.51       & 0.70        & 0.57         & 0.25         \\
WIMP~\cite{Khandelwal2020WhatIfMP}   & 1.84      & 1.11       & 0.55        & 0.43         & 0.04         \\
CXX~\cite{cxx}    & 1.63      & 1.29       & 0.69        & 0.60         & 0.08         \\
P2T~\cite{ptp}    & 1.45      & 1.16       & 0.64        & 0.46         & 0.03         \\
THOMAS~\cite{gilles2022thomas} & 1.33      & 1.04       & 0.55        & 0.42         & 0.03         \\
 &
  1.30 &
  1.00 &
  0.61 &
  0.37 &
  0.03 \\
\multirow{-2}{*}{PGP~\cite{deo2021multimodal}}   & $1.28^\dag$      & $0.95^\dag$    & $0.52^\dag$        & $0.34^\dag$        & $0.03^\dag$         \\ \midrule
XHGP  & 1.27    & 0.94       & 0.53        & 0.34         & 0.03         \\ \bottomrule
\end{tabular}
\end{table}   
 
\begin{figure}[t]
    \centering
    {\includegraphics[width=0.5\linewidth]{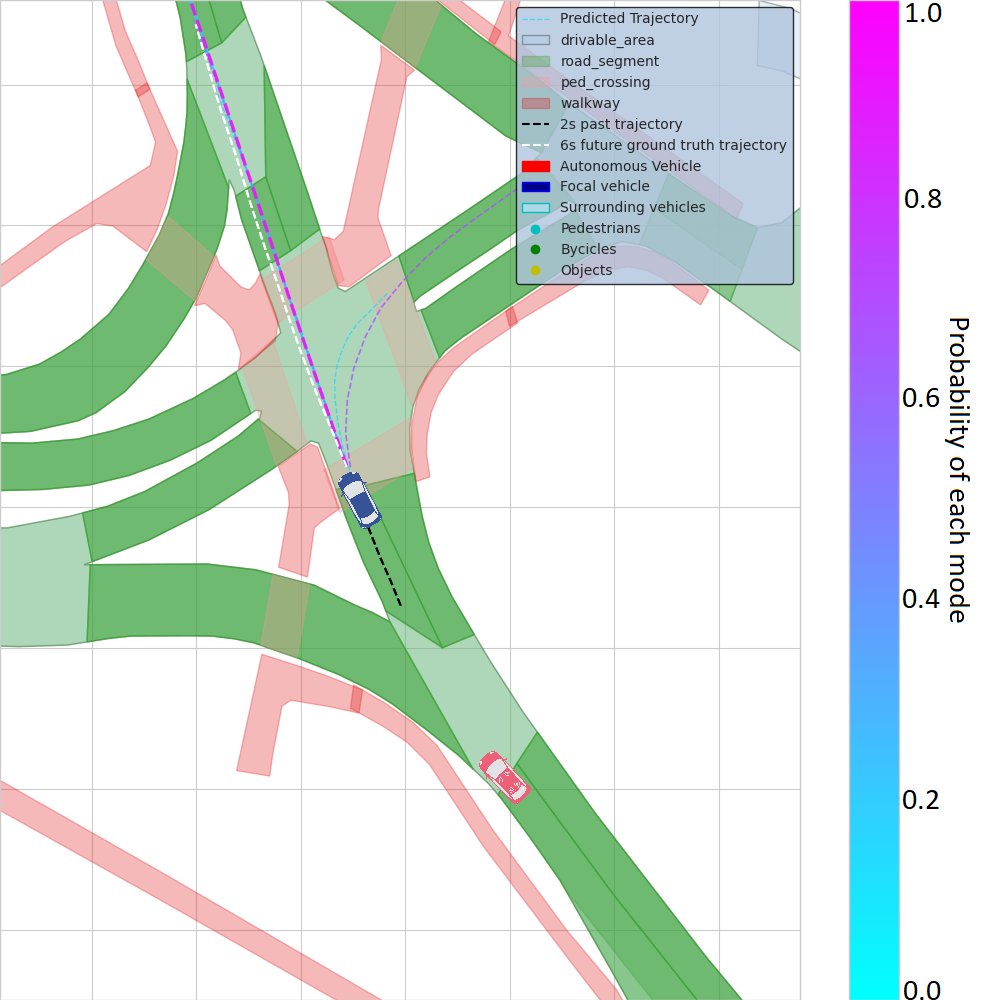}}
    \caption{Multi-modal motion prediction output representation.}
    \label{fig:mp}
\end{figure}

Figure~\ref{fig:mp} shows an example predicted scene. Two seconds of history are represented by a dashed black  line. Six seconds of future by a white dashed line. The AV is shown in red and the focal agent in dark blue. Surrounded vehicles are represented in light blue,  bicycles and motorcycles in green dots, and pedestrians in blue dots. 
Movable objects, such as traffic cones, are painted in yellow.
Ten different predictions -- modes -- are visible in the scene. The probability of each prediction is assigned to the color bar on the right. Therefore, each prediction is represented with a color and a line width describing its probability. 
In this sample scene, the most probable prediction goes straight forward. Two less likely modes turn right. The rest of the variability lies in the velocity profile. 

\subsection{Evaluation of the influence of interactions}
The black-box nature of neural networks raises many concerns, as the predictions they provide are difficult for humans to fully understand. 
The need of the explanations in autonomous driving stems from existing problems, established regulations and standards, along with opinions of the public. 
Traffic accidents and safety issues are the main reason for the need. Therefore, we proposed an explainable heterogeneous graph representation to be able to carefully explore it and indicate the most influential interactions between static and dynamic scene elements. In our study we focused on the explanation of graph temporal evolution and predicted modality to understand the influence of each interaction in the provided prediction. 

Since our graphs modeling the scene context are not very large and contain up to 100 nodes, we did not limit ourselves to just the neighborhood of the focal agent. Analysis conducted for the validation subset of the nuScenes dataset identified lanes nodes as the most influential. On the other hand, the most influential type of edges for predicting the most probable trajectory appears to be the ones connecting pedestrian-to-vehicle, and vehicle-to-vehicle. This finding is consistent with  one would expect from a human driver, as the ability to perform most maneuvers on the road depends on the geometry of the scenario (lanes, traffic lights, signs, etc.), and their exact performance is based on interactions between dynamic objects (pedestrians, cyclists, and other vehicles). However, we also detected a strong focus on vehicles and pedestrians who do not fully participate in the driven scenario, but merely indicate the limits of the road (sidewalk). This attention may be associated with the fact that, in our graphical representation of the scene, the lanes of the road are defined only by the centerlines.

\begin{figure*}[!t]
    \centering
    \begin{tabular}{cccc}
         {\includegraphics[trim={0 0 5cm 0},clip,width=0.31\linewidth]{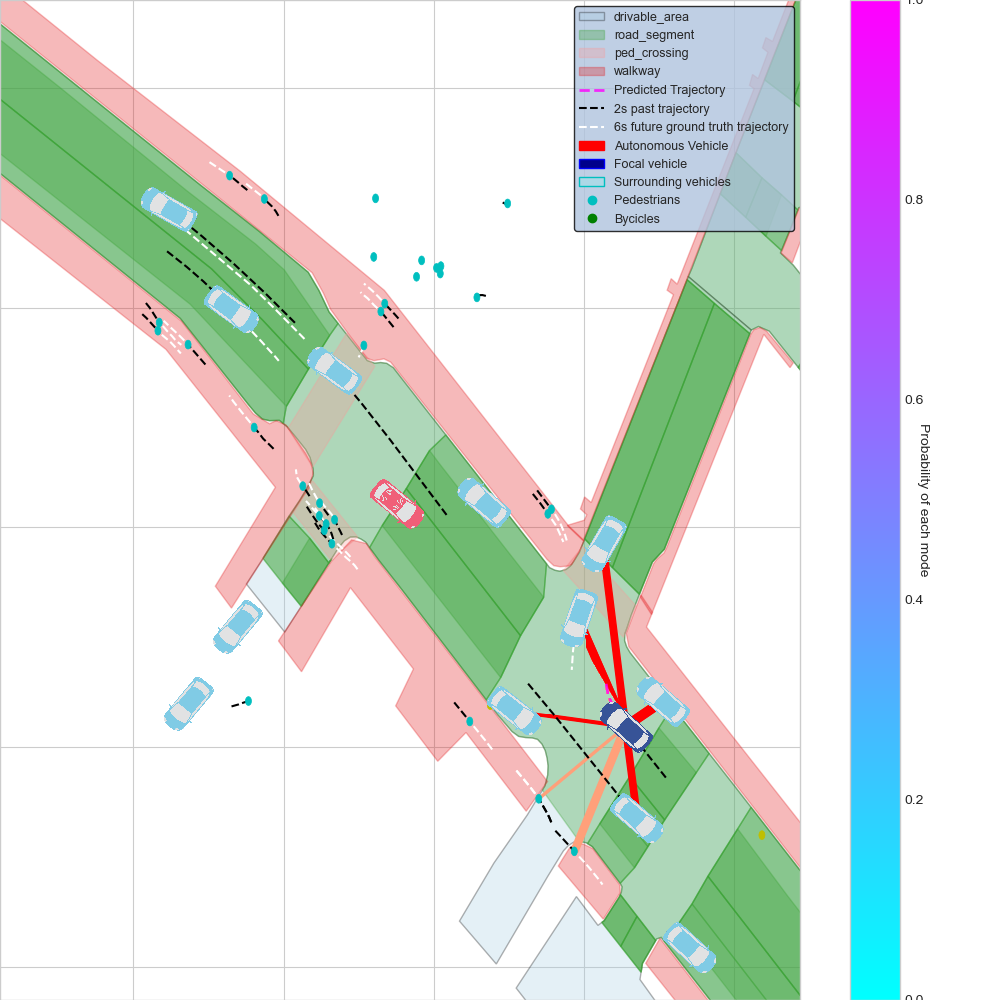}} &
         {\includegraphics[trim={0 0 5cm 0},clip,width=0.31\linewidth]{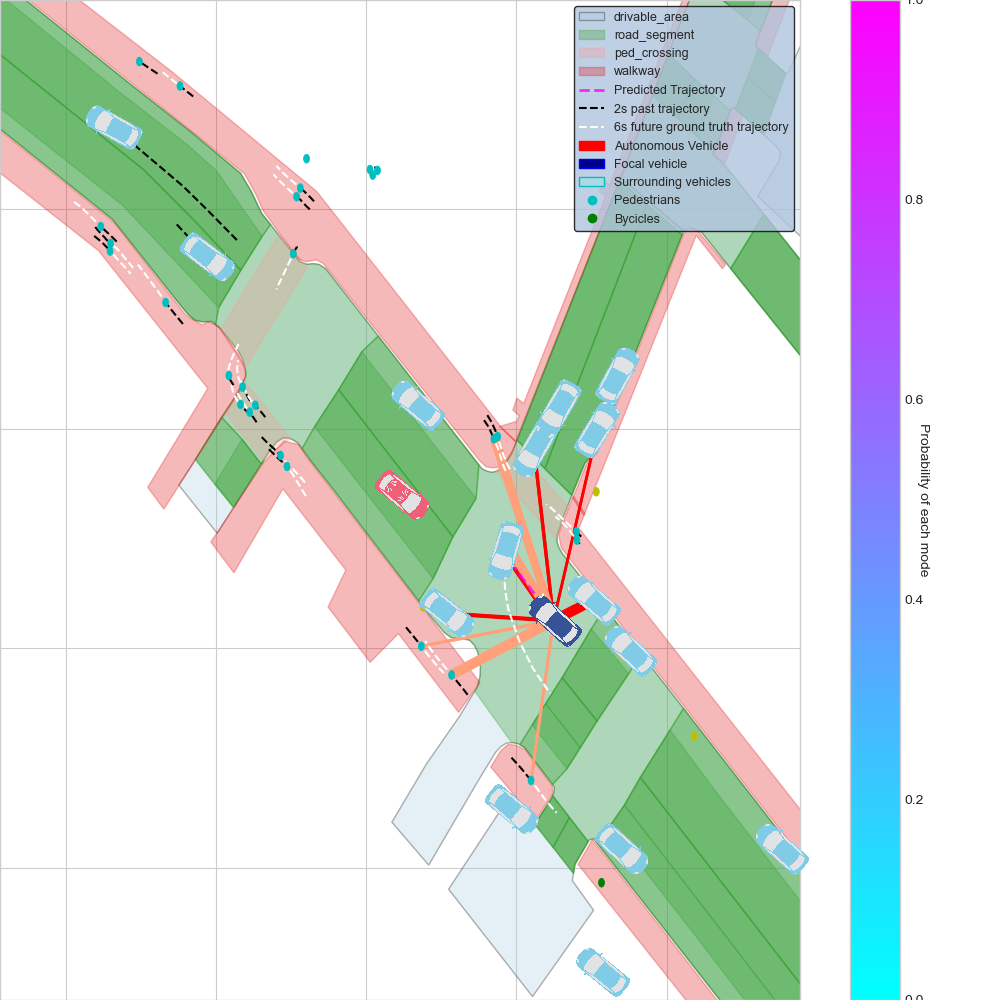}} &
         {\includegraphics[trim={0 0 5cm 0},clip,width=0.31\linewidth]{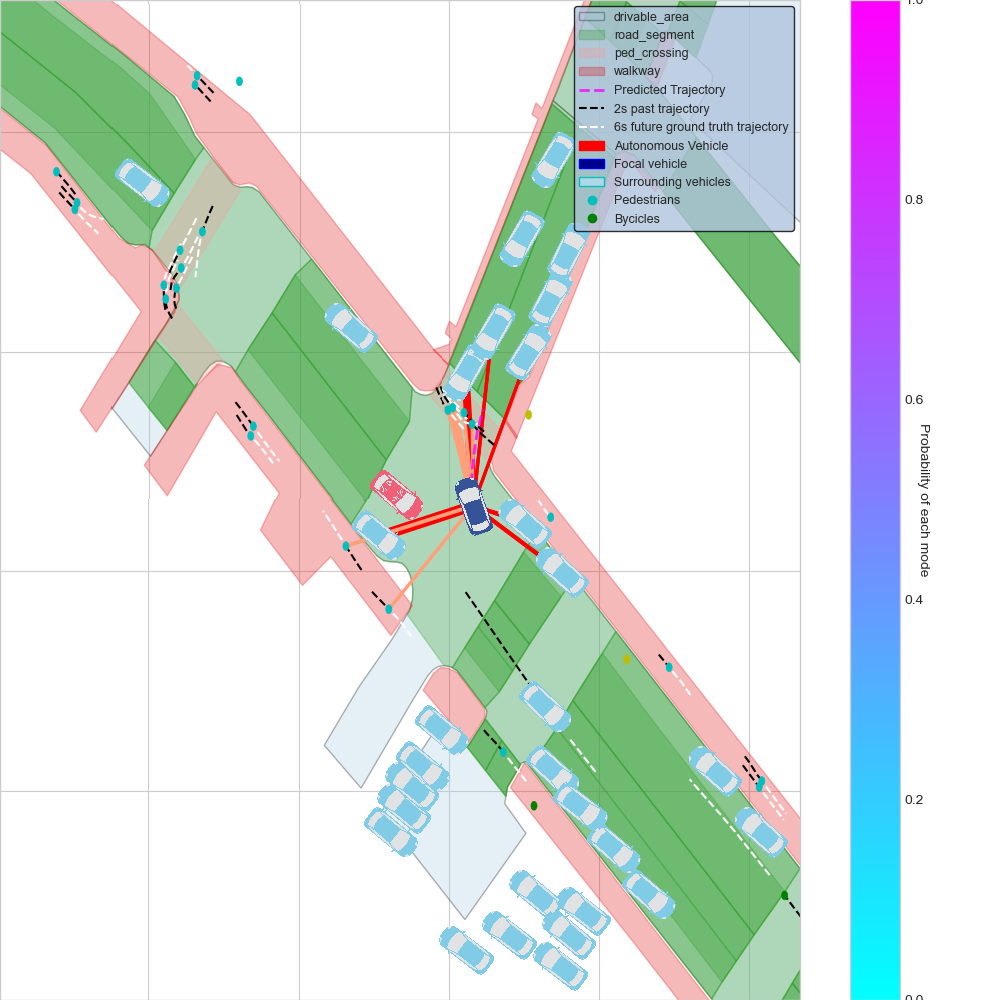}}
    \end{tabular}
    \begin{tabular}{cccc}
         {\includegraphics[trim={0 0 5cm 0},clip,width=0.31\linewidth]{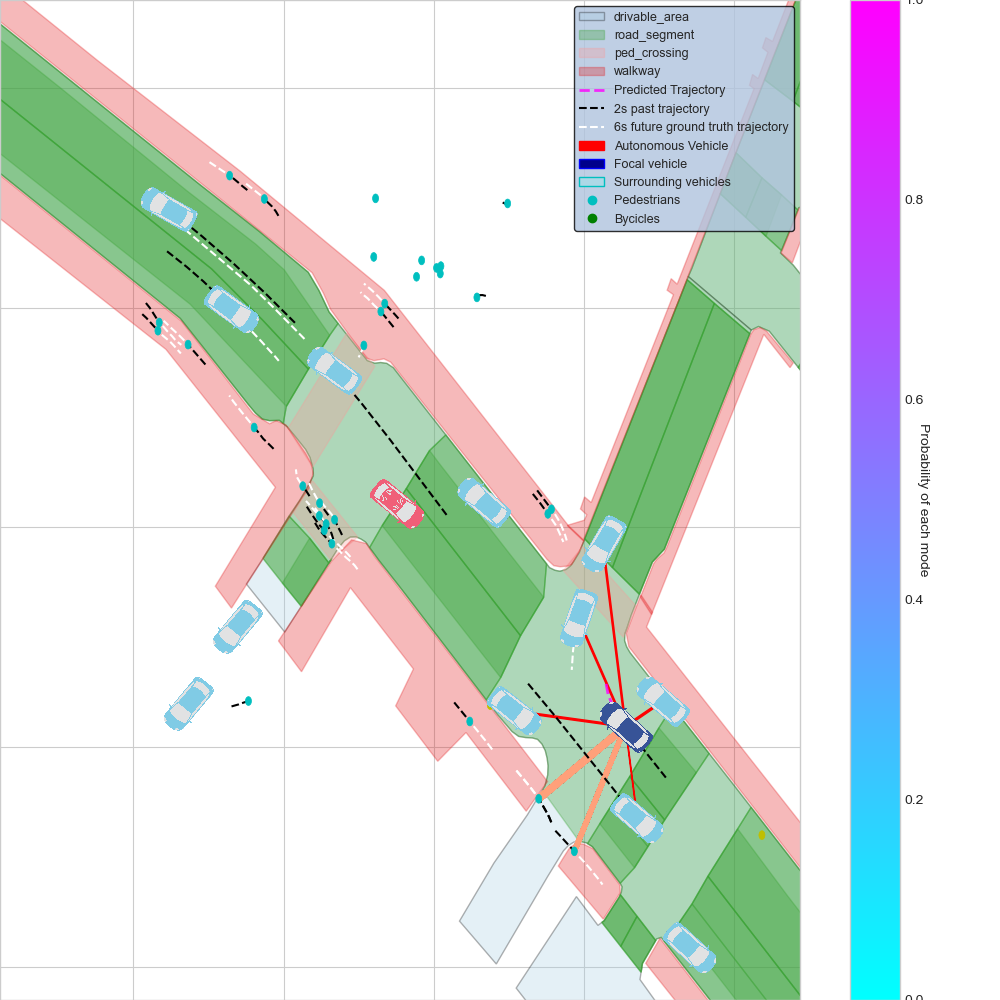}} &
         {\includegraphics[trim={0 0 5cm 0},clip,width=0.31\linewidth]{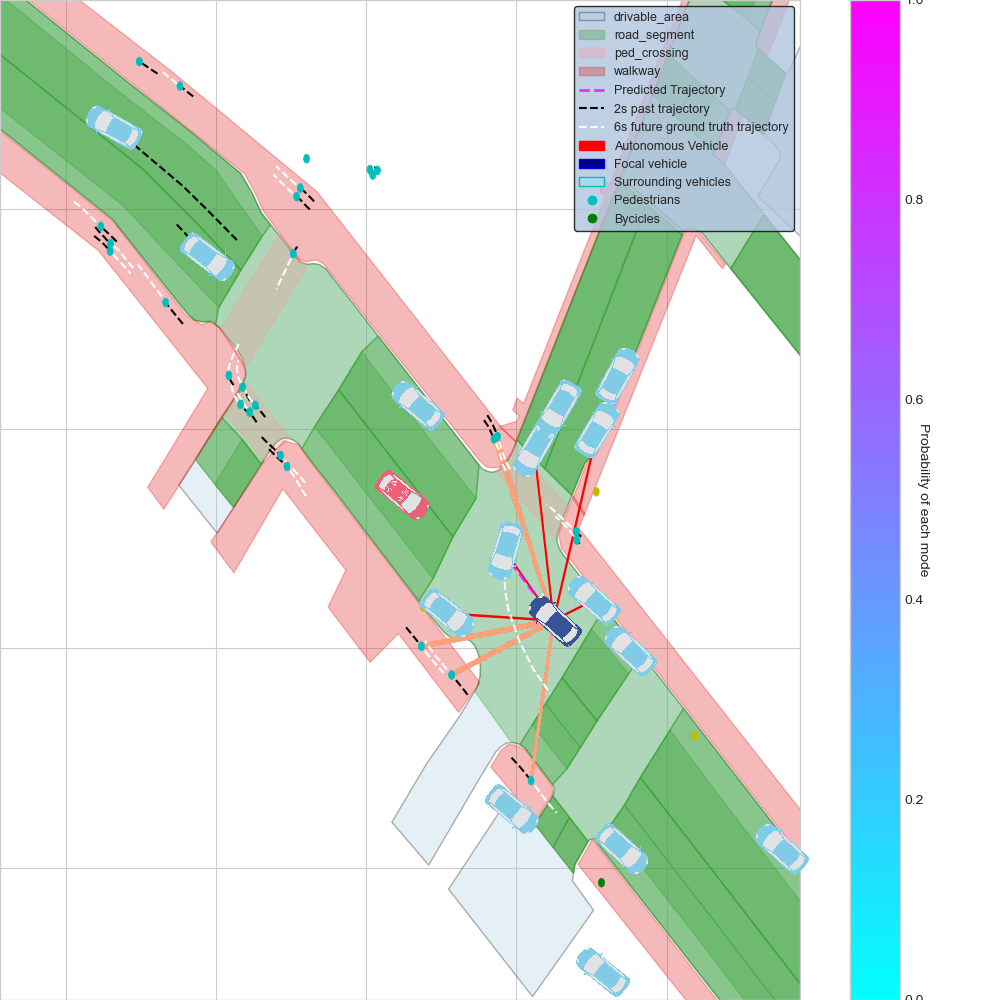}} & 
         {\includegraphics[trim={0 0 5cm 0},clip,width=0.31\linewidth]{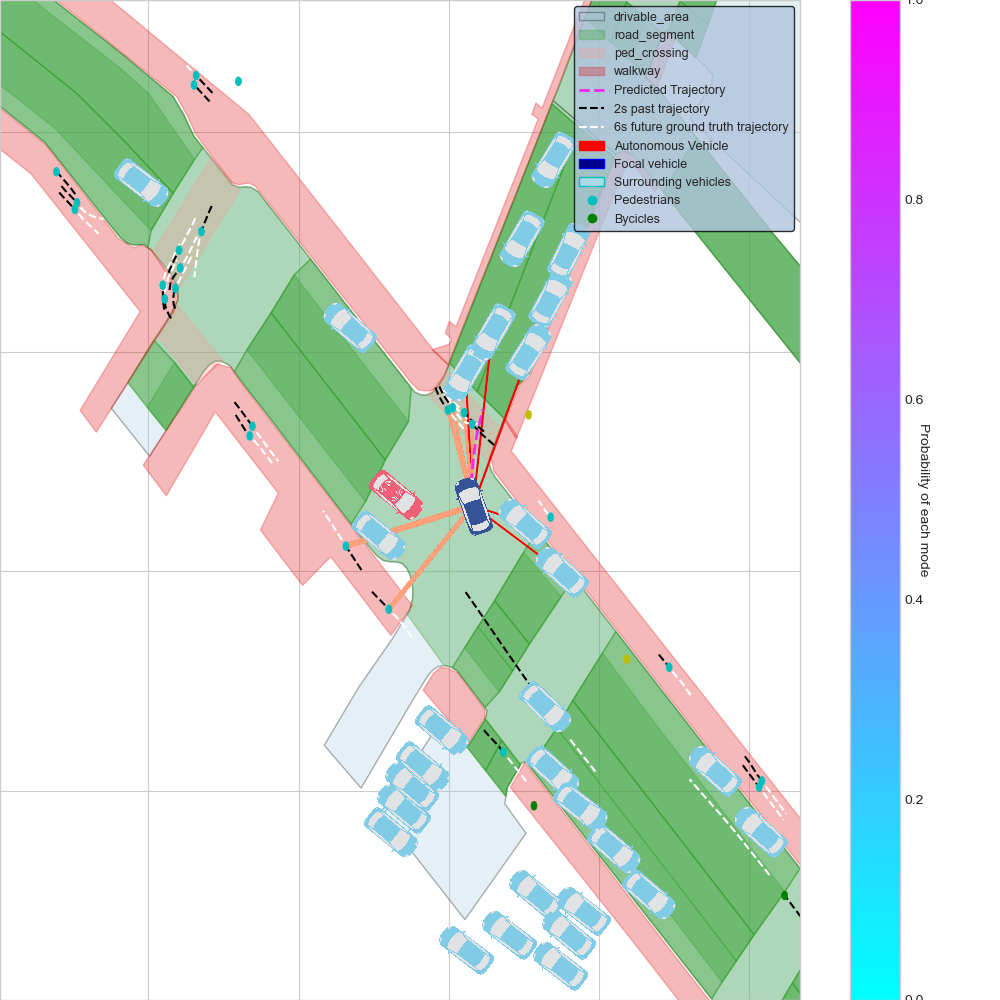}}
    \end{tabular}
    \caption{\textbf{Qualitative results}: importance of interactions for vehicle-to-vehicle (red) and pedestrian-to-vehicle (pink) with the focal agent based on GNNExplainer (\textit{top row}) and learned attention (\textit{bottom row}).} 
    \label{fig:temporal}
\end{figure*}

The influence of interactions was first visualised for the most likely trajectory predicted by our model. To do so, we focus on selected scenes, trying to understand the basic rules for these predictions. 
Given the findings stated in the previous paragraph, we focus on the interactions with dynamic agents. 
Figure~\ref{fig:temporal} depicts explanations, i.e., interaction importance, for a city center scenario with a large number of pedestrians and vehicles -- both parked and in movement. The focal agent continues its trajectory in a straight line entering an intersection  with the intention of turning right, stopping at the pedestrian crossing. The Figure presents the most important vehicle-to-vehicle and pedestrian-to-vehicle interactions for the focal agent. Top row indicates the masks learnt by GNNExplainer, while bottom row depicts the learnt attention weight (marked in red -- interactions with vehicles, and in pink -- interactions with pedestrians) between each node connected with the focal agent. Each column represents a different timestamp for the given scenario, moving from left to right. The thickness of the edges represents the importance level of each interaction, which changes for each frame. Overall, GNNExplainer and the visualized attention indicate similar interactions as the important ones, however GNNExplainer appears to be more sensitive for all interactions with focal agent.

\begin{figure*}[!t]
    \centering
     {\includegraphics[trim={0 0 5cm 0},clip,width=0.3\linewidth]{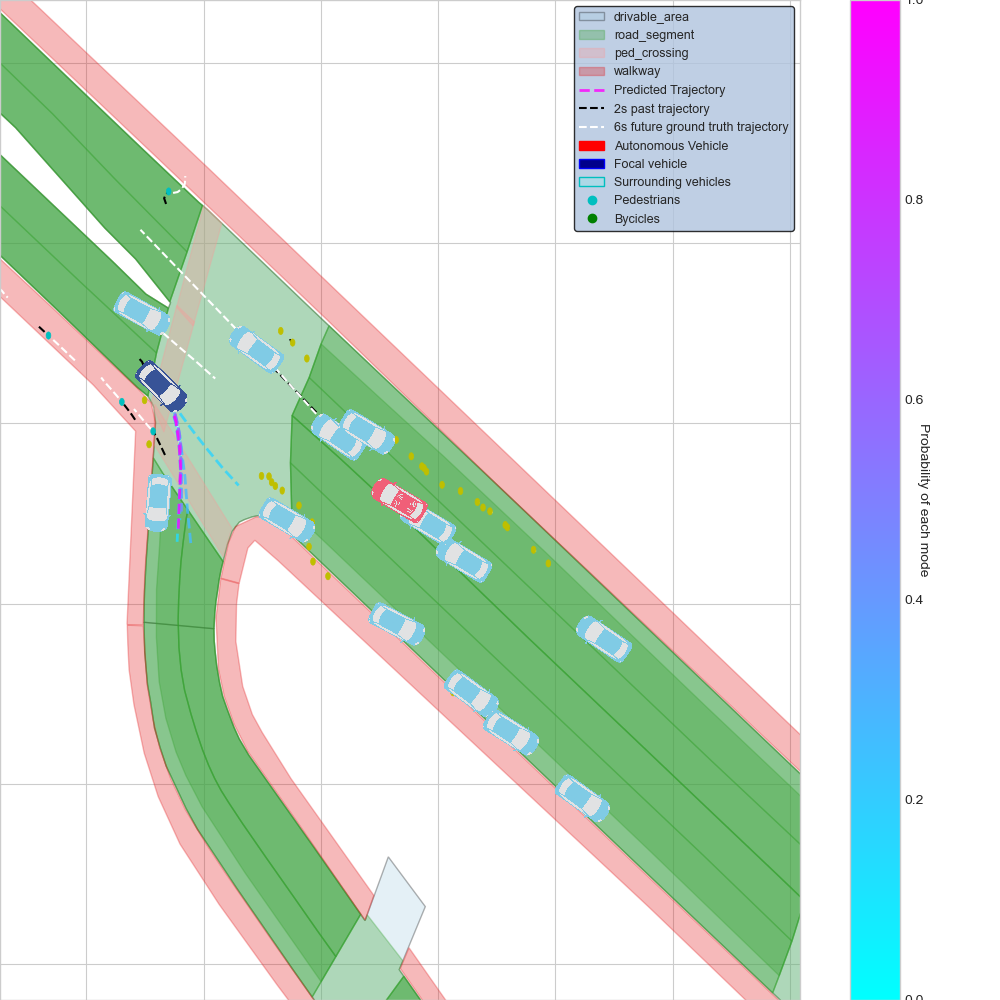}} 
     {\includegraphics[trim={0 0 5cm 0},clip,width=0.3\linewidth]{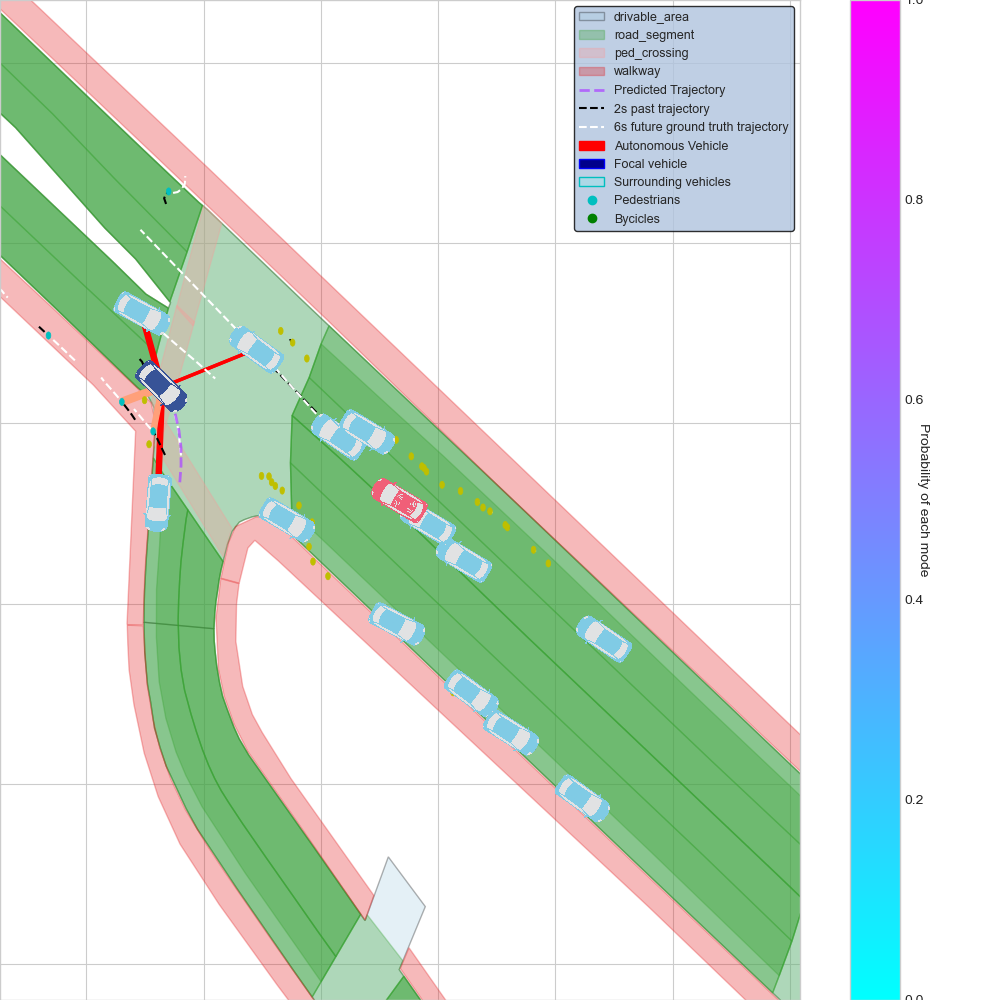} }
     {\includegraphics[trim={0 0 5mm 0},clip,width=0.367\linewidth]{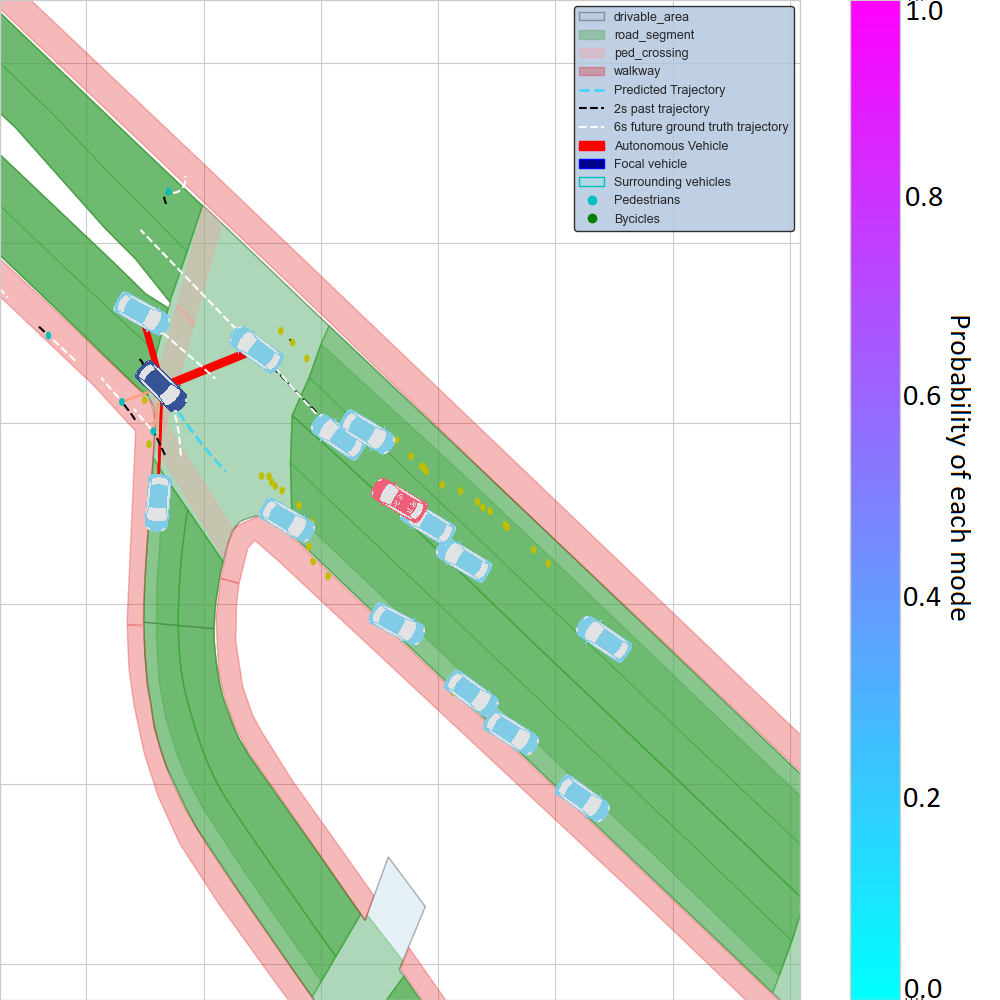} } 
     \caption{\textbf{Qualitative results}: importance of interactions for vehicle-to-vehicle (red) and pedestrian-to-vehicle (pink) with the focal agent based on GNNExplainer for different modalities. The first image shows the 10 possible modes predicted by the model indicating 2 different maneuvers -- turning right and going straight -- explained in the next images.}
    \label{fig:modality}
\end{figure*}

The results explained in terms of interaction importance can be produced also for different modalities, not only the most probable trajectory. Figure~\ref{fig:modality} shows a scenario with the focal agent on a T-junction, attempting to turn right. The first image shows the ten possible modalities predicted by the model, the next images show the importance of interactions between agents for the selected predicted trajectory. We can observe two different behaviors: turning right (most likely prediction, coinciding with the ground-truth), and continuing straight (less likely). In the former, the interaction between the central agent and pedestrians crossing the road was identified as the most important interaction. In this situation, the car slowly entered the intersection and turned right. For the other mode, a straight trajectory is observed with an increase in the importance of the interactions with surrounding cars. In this case, the interaction with pedestrians on the sidewalk -- which indicates the roadway boundary -- was of secondary importance. 

\begin{table}[!t]
\centering
\caption{Performance evaluation of GNNExplainer and attention baseline explainability approaches.}
\label{tab:exp}
\begin{tabular}{@{}c|ccc@{}}
\toprule
Method       & \textit{Fidelity+} & \textit{Fidelity-} & \textit{Sparsity} \\ \midrule
GNNExplainer &   0.47             &    0.97            &    0.81           \\
Attention    &   0.57             &    0.96            &    0.64           \\ \bottomrule
\end{tabular}
\end{table}

To provide a quantitative validation of the trained explainers for the most probable trajectory, we further calculated metrics described in Section~\ref{sec:gnnexp_metrics}. Note that, when the explanation results are soft values (i.e., $m_i \in [0,1]$) as it is in our the case, the \textit{Sparsity} is determined by a threshold value. Intuitively, larger threshold values tend to identify fewer features as important and, hence, increase the \textit{Sparsity} score and decrease the \textit{Fidelity+} score. The Table~\ref{tab:exp} presents results for GNNExplainer and learnt attentions for $m_i \ge 0.8$. 
The scores obtained are similar for both methods, which can also be observed in the qualitative analyses presented above. The higher \textit{Sparsity} value for GNNExplainer indicates that the explanations provided by this method are more sparse -- i.e., it tends to give higher importance for fewer input features. This is also related to the main idea of this approach of finding the minimum sub-graph.

\subsection{Counterfactual explanations}

In this section we explore a series of counterfactual experiments to explain challenging scenarios and demonstrate how the output distribution is highly influenced by the scene context. 

\begin{table*}[ht]
\centering
\caption{Ablation of lanes and dynamic agents. Results for K=10 simulating different recall levels for lane and dynamic agent detection.  }
\label{tab:robustness}
\resizebox{\textwidth}{!}{%
\begin{tabular}{|cc|ccccc|}
\hline
\textbf{Object}                 & \textbf{Experiment}                                                   & \textbf{minADE\_5} & \textbf{minADE\_10} & \textbf{miss rate\_5} & \textbf{miss rate\_10} & \textbf{BC} \\ \hline
                                & \begin{tabular}[c]{@{}c@{}}Baseline\\ Recall 100\%, K=10\end{tabular} & 1.27               & 0.94                & 0.53                  & 0.34                   & 1.85        \\ \hline
\multirow{6}{*}{Lanes}          & Recall 95\%                                                           & 1.36               & 1.00                & 0.54                  & 0.36                   & 2.18        \\
                                & Recall 90\%                                                           & 1.44               & 1.05                & 0.57                  & 0.39                   & 2.47        \\
                                & Recall 80\%                                                           & 1.61               & 1.16                & 0.61                  & 0.43                   & 3.08        \\
                                & Recall 50\%                                                           & 2.11               & 1.51                & 0.70                  & 0.55                   & 3.98        \\
                                & Recall 20\%                                                           & 2.49               & 1.83                & 0.76                  & 0.62                   & 4.07        \\
                                & Recall 0\%                                                            & 2.72               & 2.11                & 0.80                  & 0.64                   & 3.86        \\ \hline
\multirow{6}{*}{Dynamic agents} & Recall 95\%                                                           & 1.30               & 0.97                & 0.52                  & 0.34                   & 1.93        \\
                                & Recall 90\%                                                           & 1.31               & 0.97                & 0.53                  & 0.35                   & 1.93        \\
                                & Recall 80\%                                                           & 1.32               & 0.97                & 0.53                  & 0.34                   & 1.94        \\
                                & Recall 50\%                                                           & 1.34               & 0.98                & 0.54                  & 0.36                   & 1.96        \\
                                & Recall 20\%                                                           & 1.38               & 0.99                & 0.57                  & 0.37                   & 1.91        \\
                                & Recall 0\%                                                            & 1.38               & 0.99                & 0.58                  & 0.37                   & 1.96        \\ \hline
\end{tabular}%
}
\end{table*}

\begin{figure*}[!t]
    \centering
    \begin{tabular}{ccc}
         {\includegraphics[trim={0 0 5cm 0},clip,width=0.3\linewidth]{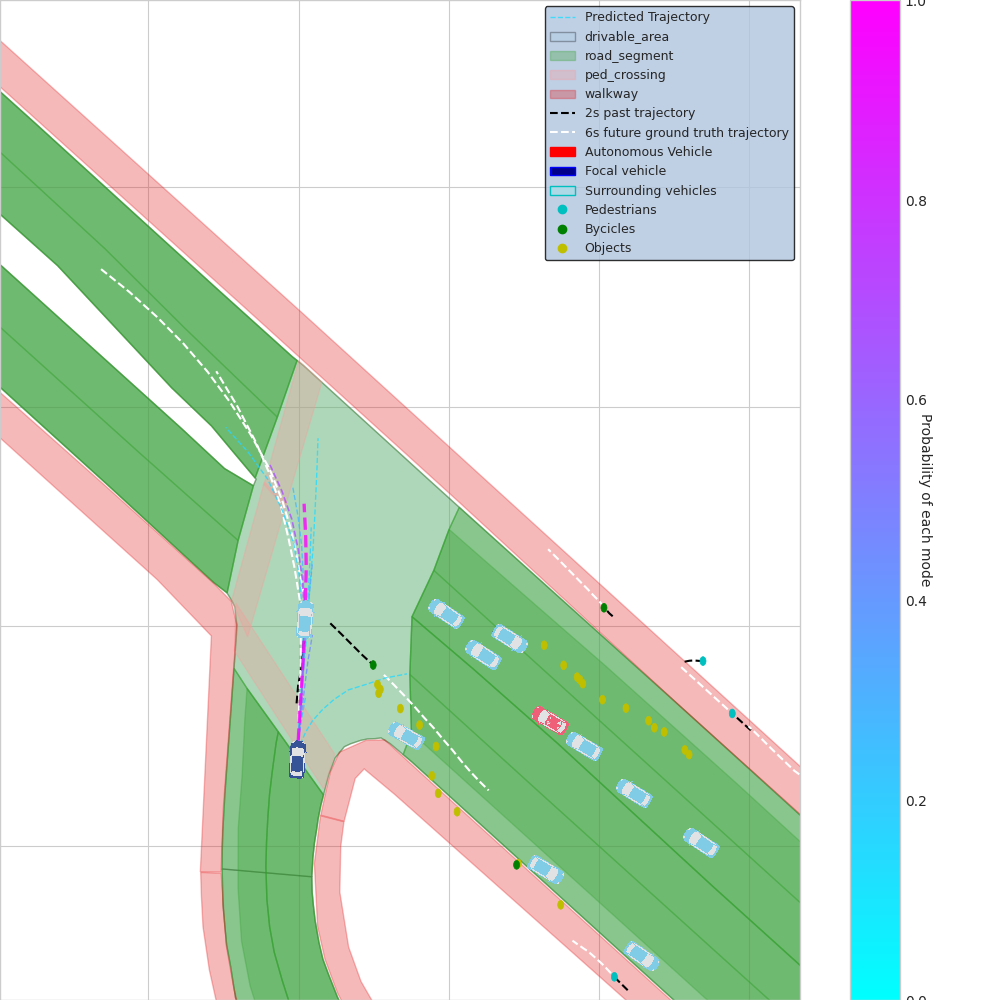} } 
         {\includegraphics[trim={0 0 5cm 0},clip,width=0.3\linewidth]{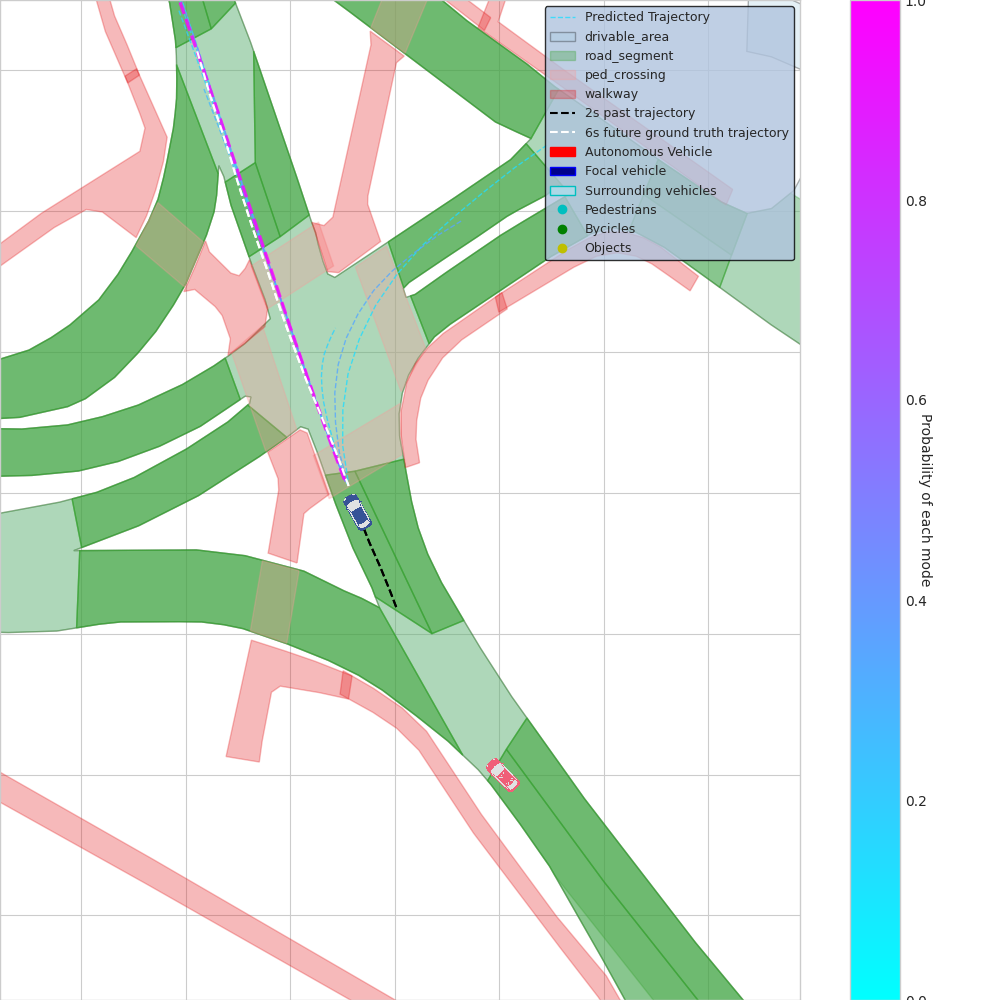}}
         {\includegraphics[trim={0 0 5cm 0},clip,width=0.3\linewidth]{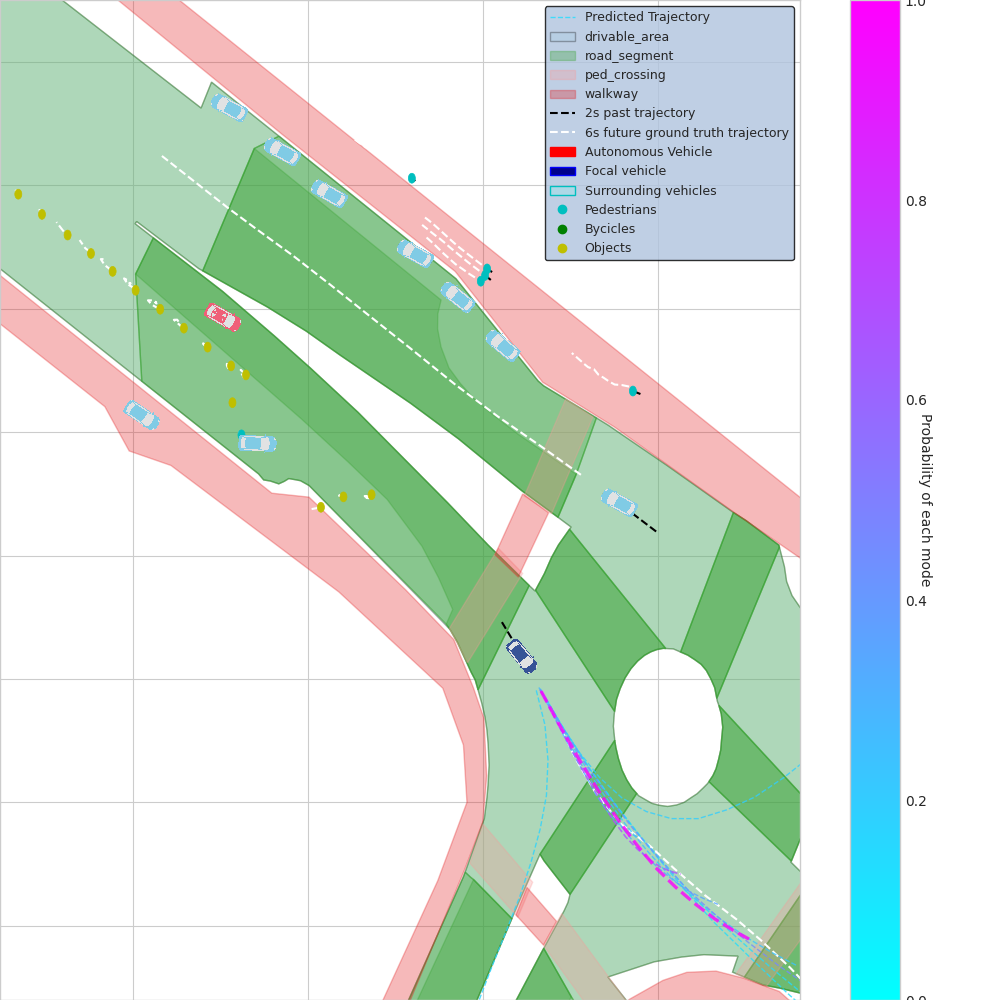}}
    \end{tabular}
    \begin{tabular}{ccc}
         {\includegraphics[trim={0 0 5cm 0},clip,width=0.3\linewidth]{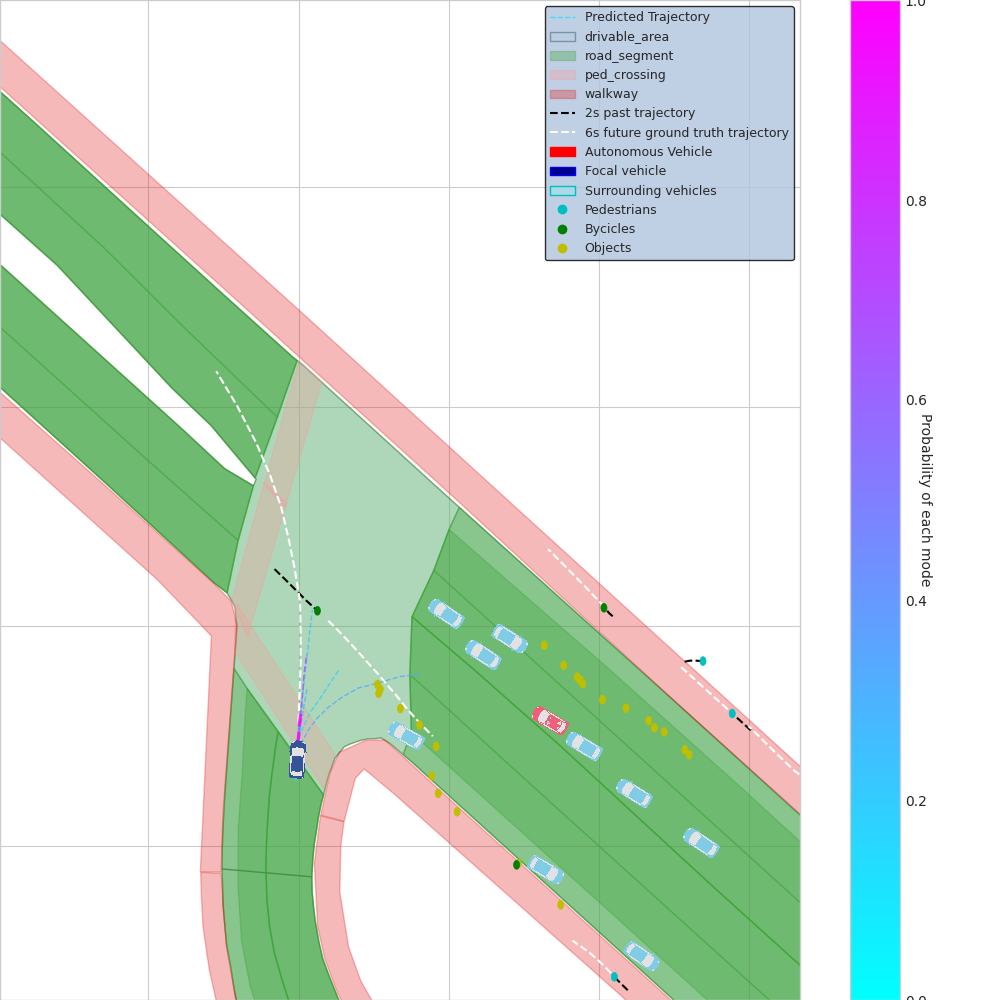} } 
         {\includegraphics[trim={0 0 5cm 0},clip,width=0.3\linewidth]{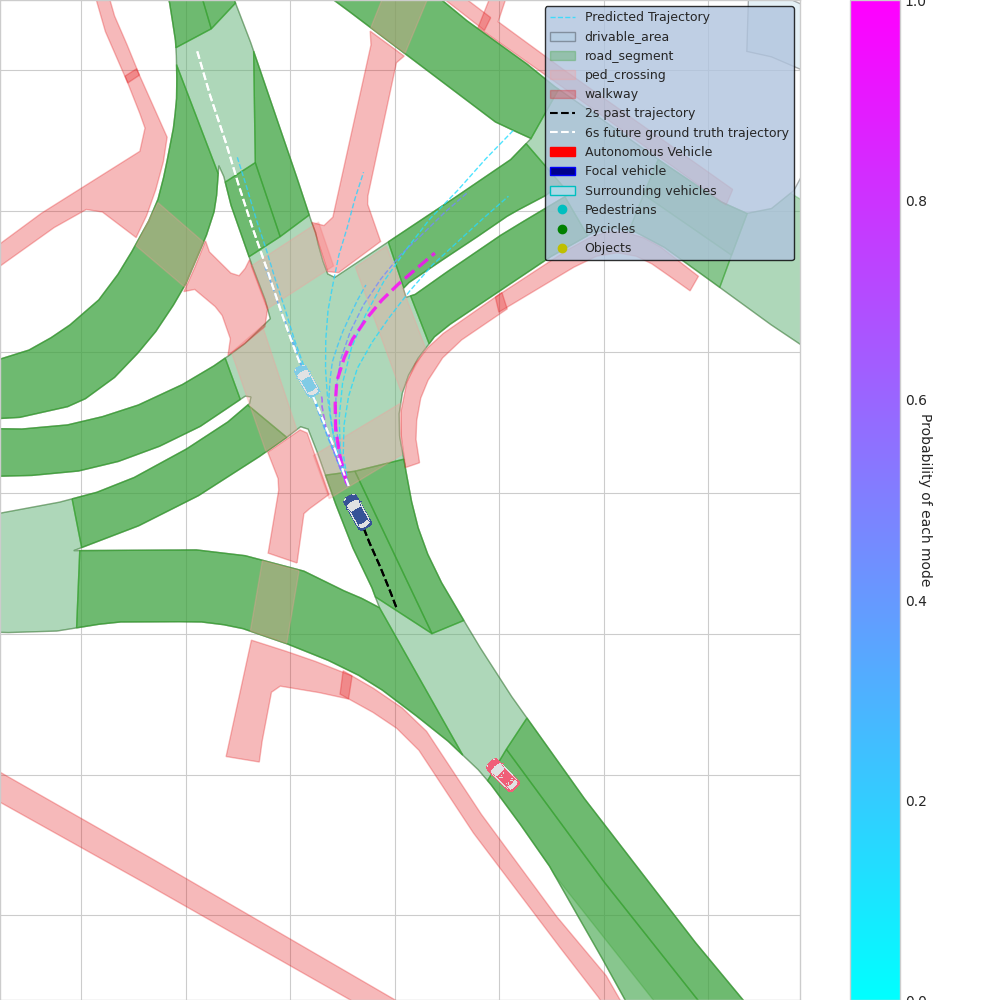} }
         {\includegraphics[trim={0 0 5cm 0},clip,width=0.3\linewidth]{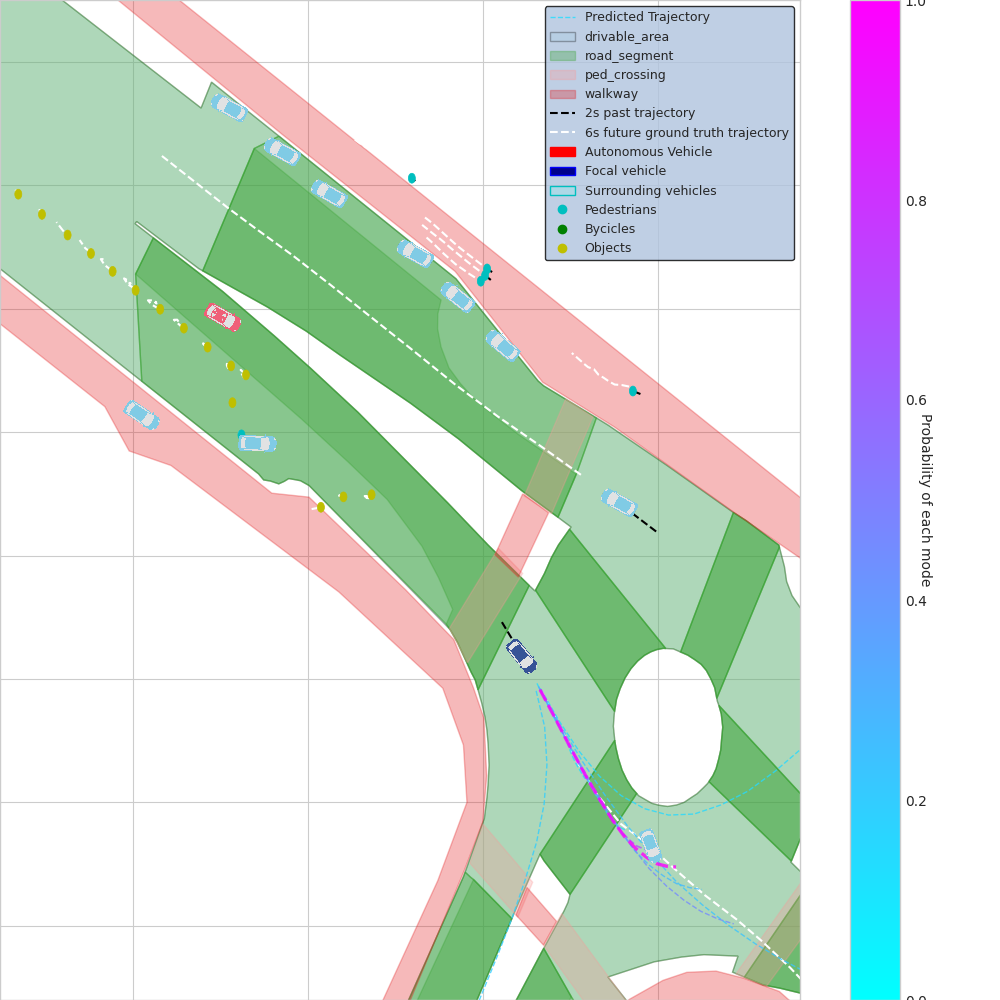}}
    \end{tabular}
    \caption{Qualitative results of counterfactual insertion. Top row shows the real scene with ten predicted modes. Bottom row shows same scene with the counterfactual and the modified predictions. In the first scenario, velocity is highly reduced to avoid the collision with the bicycle, which now intersects the ground-truth trajectory. In the second scenario, the most likely predictions turn right to avoid the vehicle stopped in front. In the last scenario, the velocity is reduced to avoid a collision with the vehicle in the middle of the roundabout. } 
    \label{fig:counterfactuals}
\end{figure*}

\textbf{Agent insertion.}
Figure \ref{fig:counterfactuals} shows four scenarios where fictitious agents are inserted in the scene or behaviors of current agents are modified.

In the first scene, the trajectory of the bicycle is modified so that it intersects with the focal agent ground-truth trajectory. The vehicle in front of the focal agent is removed. Two predictions appear turning right, whereas the other eight predictions reduce drastically their velocity to allow the bicycle to safely cross the intersection. 
If the vehicle in front is not removed from the scene, the predictions appear to be less conservative and follow the vehicle trajectory. However, velocity of the predictions are still reduced and two modes appear turning right. 
In the second scenario, we insert a  fictitious vehicle stopped in front of the focal agent. In this case, most of the predictions change, turning right instead of following a straight path. However, two  predictions with low probability go off the road. The least probable prediction continues straight. We believe that this mode tries to cover the scenario in which the stopped vehicle continues its trajectory. Nonetheless, these low probability predictions should be taken into account with caution in order to plan a safe maneuver.  
In the last scenario, a vehicle is inserted in the middle of the roundabout. The focal vehicle speed is reduced to follow this vehicle. Yet, some modes seem to try to overtake it.

These counterfactual examples show that the model produces sensible output predictions when encountering new unexpected scenarios. However, given that we predict ten modes and seek to cover the underlying output distribution, we observe some low-probability modes that adopt less cautious and more reckless behaviors. The decision making system should take into consideration all probable scenarios with its associated likelihood to plan a safe and efficient trajectory. 

\begin{figure}[!t]
    \centering
    {\includegraphics[width=0.7\linewidth]{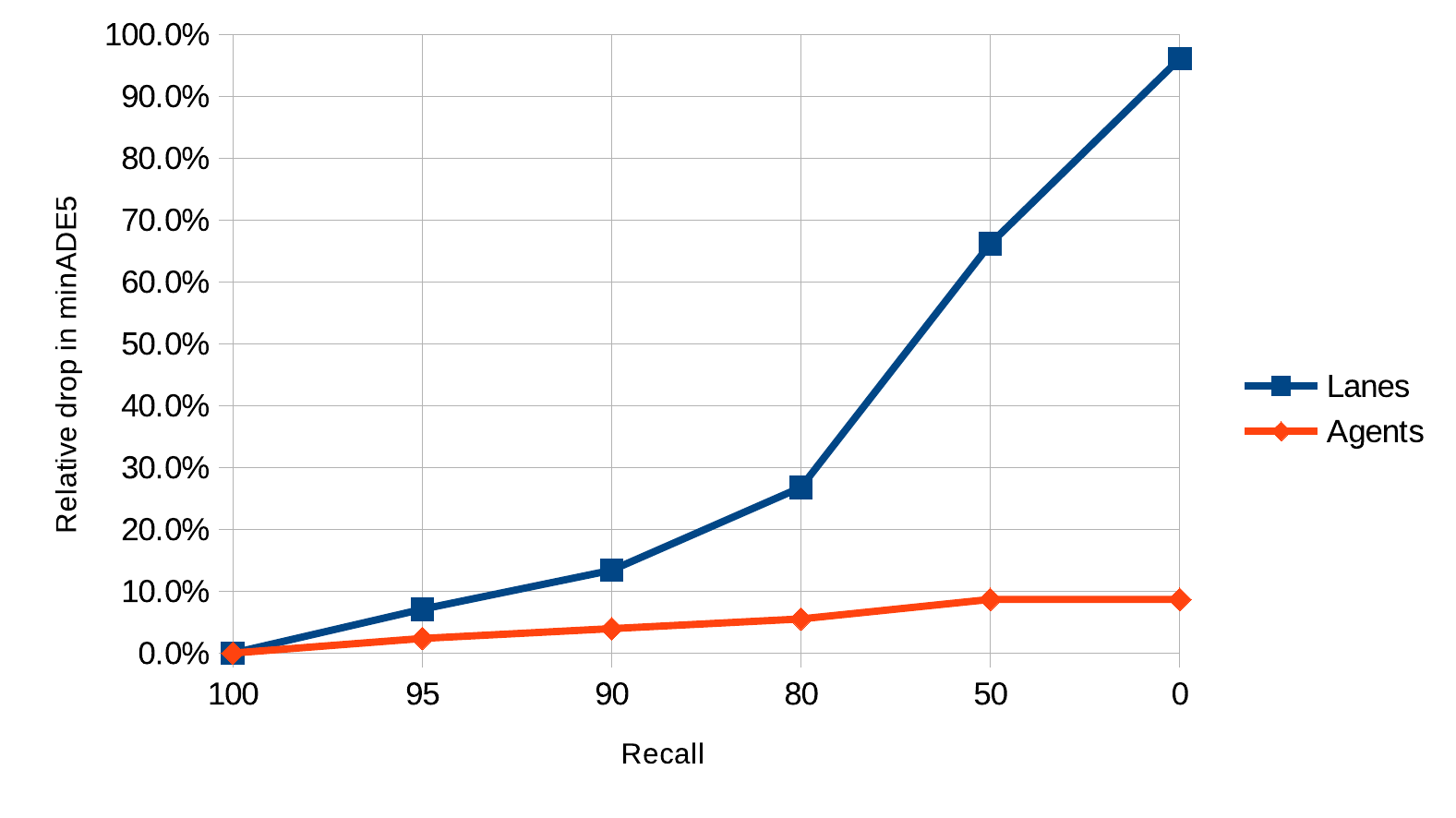}}
    \caption{Relative drop in $minADE_5$ under different recall levels for lane and dynamic agents detection.}
    \label{fig:ablation}
\end{figure}

\textbf{Object ablation.}
In addition, we perform a qualitative and quantitative evaluation of the effect of lanes and dynamic agents on the output. 
Lanes insert important inductive biases in the model and are, hence, crucial for its performance. On the other hand, the ability to capture interactions is essential for a safe and efficient planning.  
We perform an ablation in an independent manner for lanes and dynamic agents, simulating noise in the perception system.
Table \ref{tab:robustness} shows the performance of the model as having an object detector with a recall from 95\% -- which is close to the demand for self-driving cars -- to 0\%, i.e. no lanes or dynamic agents detected at all. Figure \ref{fig:ablation} shows the relative evolution of $minADE_5$ in a graphical manner.

When the recall is 0\% for lane detection, the system can only rely on the focal agent's past trajectory and other dynamic agents, with no information about the road topology.
In case of masking all agents in the scene, we evaluate the impact on performance when interactions are not taking into account for the prediction. 
For the ablation of dynamic agents, the noise is introduced in the temporal dimension, meaning that we mask random frames of the agents that interact with the focal agent. 

Performance is noticeably reduced as the recall of the object detector decreases. The most detrimental effect is shown for lane masking. These results suggest that the model relies heavily on lane information to make  its predictions given the significant inductive biases they introduce. However, the results are still sensible when no lanes are detected. Interactions also prove to be an important factor in model performance, albeit in a much lower scale. This is probably due to the fact that most driving trajectories can be inferred solely from the agent's past trajectory and road topology. In addition, the data do not include enough scenes in which interactions are crucial for the behavior of the focal agent. Nevertheless, information from other dynamic agents in the scene is essential to produce more sensible predictions, avoiding collisions and dangerous behaviors that are not consistent with the traffic scene.  

In order to better understand the behavior of the system under noise conditions, we make a qualitative evaluation ablating specific important lanes. More concretely, we mask lanes traveled by the ground-truth trajectory. For all experiments, we evaluate the behavior when only some lane segments in the trajectory are masked, as well as when masking the entire ground-truth path. When only one lane segment is removed from the input, the model behavior is practically the same. However, when more lane segments are masked, predictions change to find an alternative plausible route. 
  
\begin{figure*}[!t]
    \centering
    \begin{tabular}{cccc}
         {\includegraphics[trim={0 0 5cm 0},clip,width=0.23\linewidth]{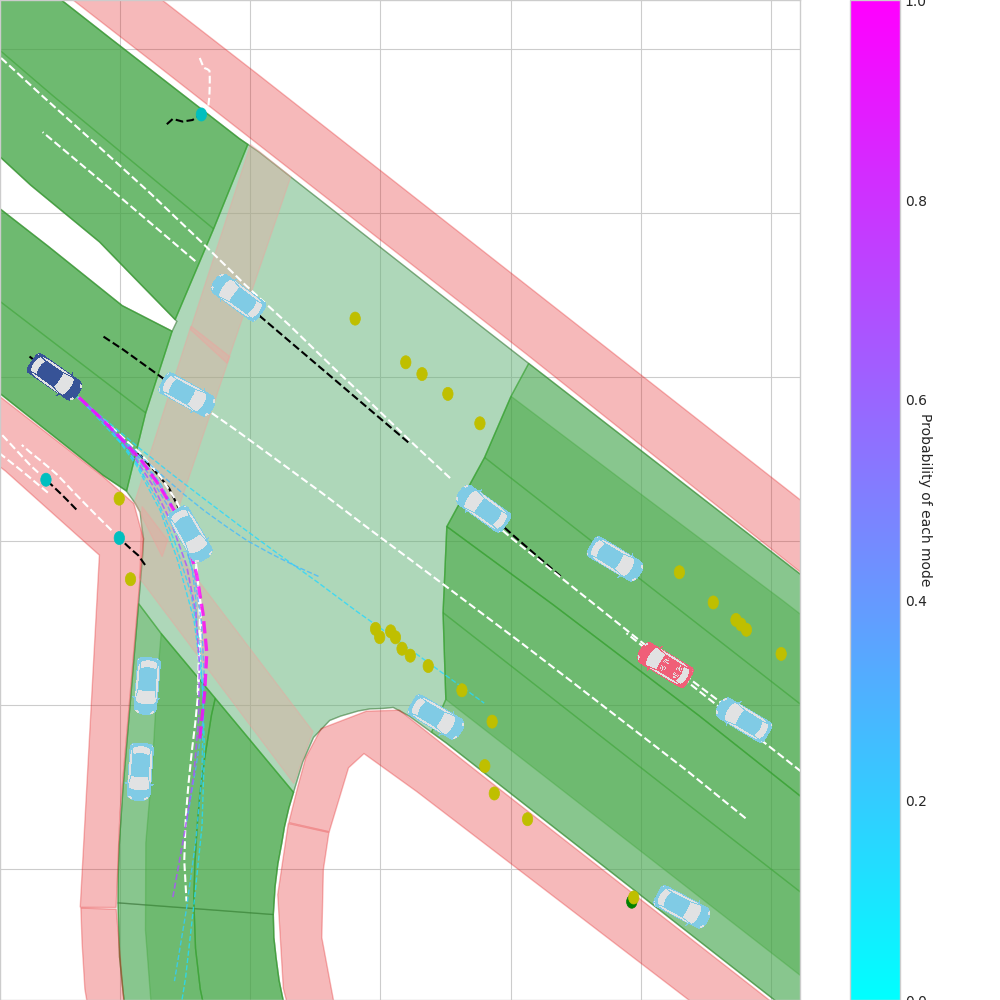} } 
         {\includegraphics[trim={0 0 5cm 0},clip,width=0.23\linewidth]{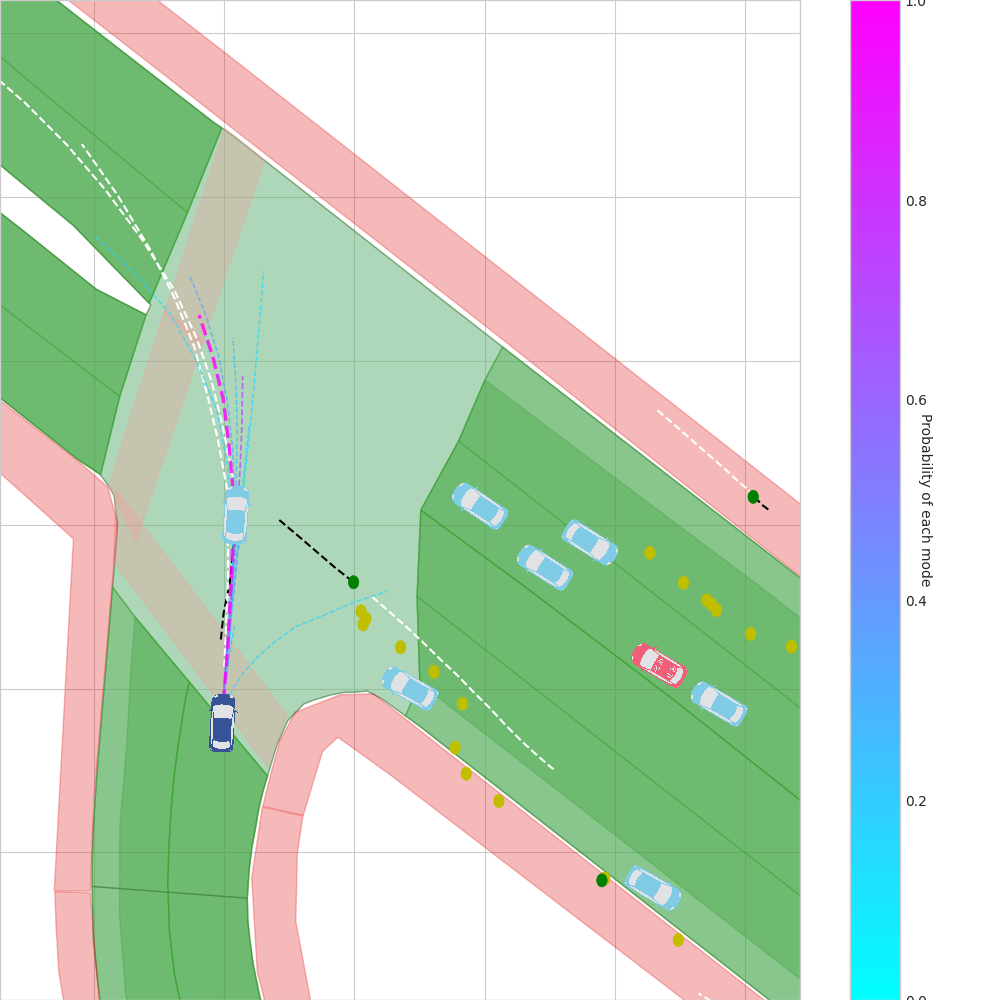}}
         {\includegraphics[trim={0 0 5cm 0},clip,width=0.23\linewidth]{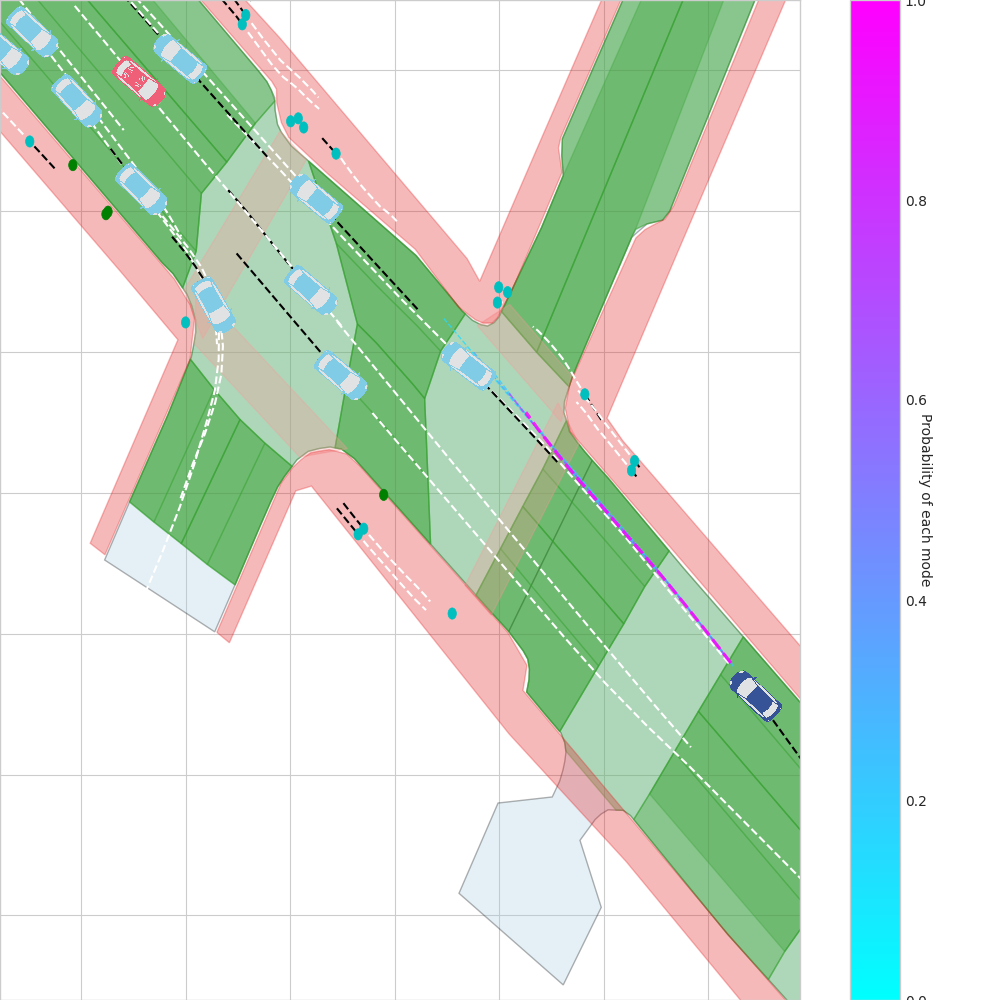} }
         {\includegraphics[trim={0 0 5cm 0},clip,width=0.23\linewidth]{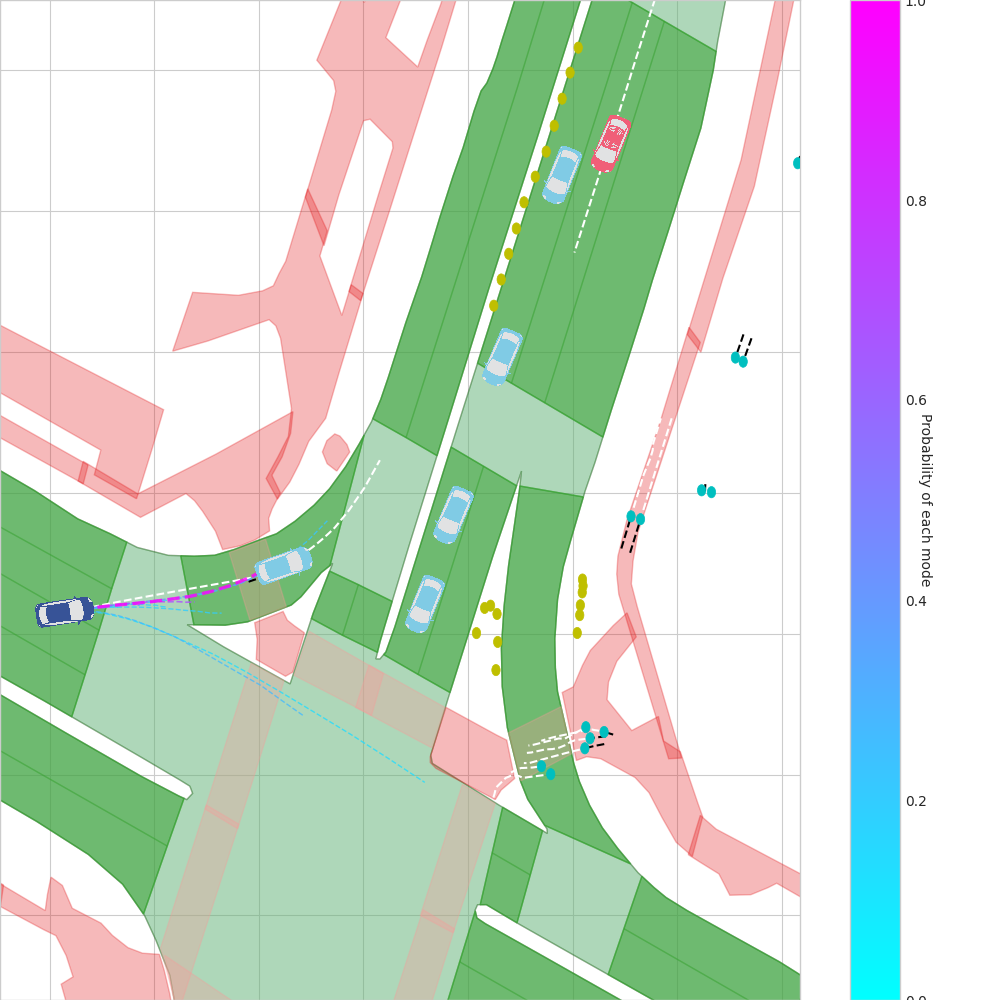}}
    \end{tabular}
    \begin{tabular}{cccc}
         {\includegraphics[trim={0 0 5cm 0},clip,width=0.23\linewidth]{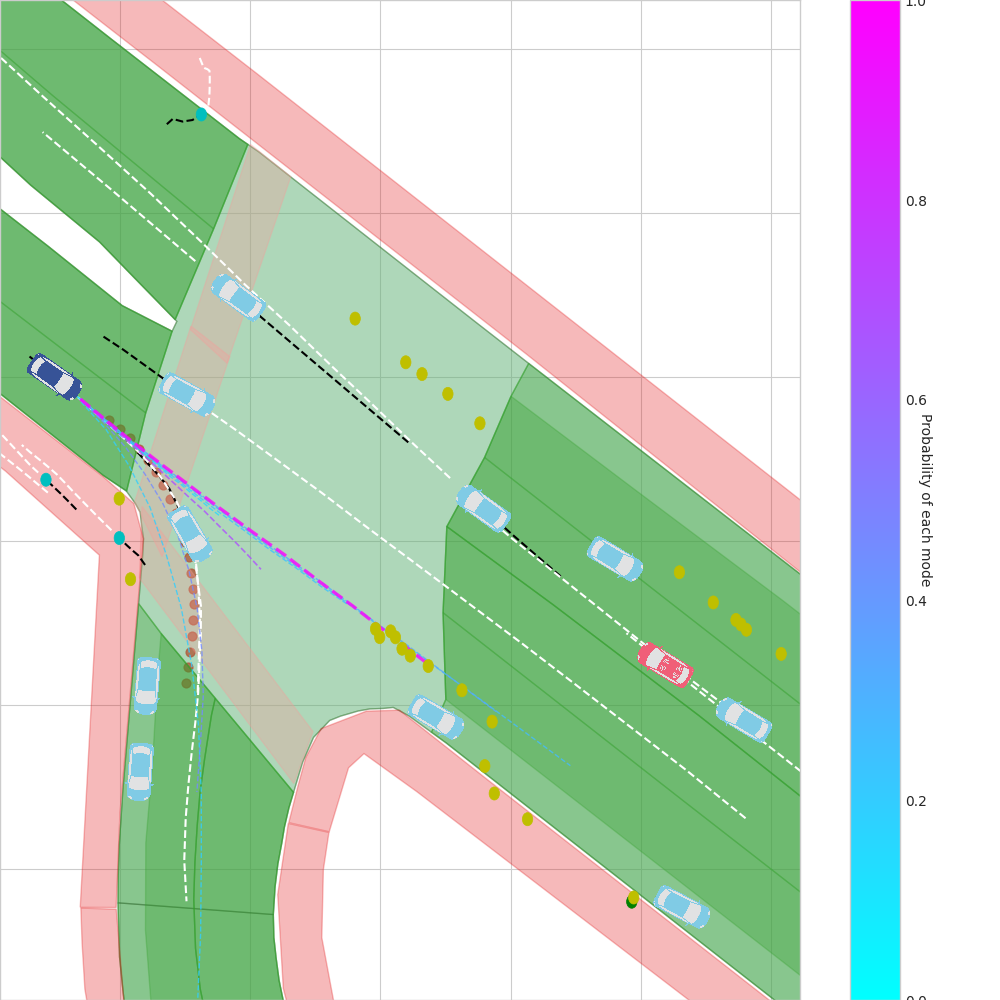} } 
         {\includegraphics[trim={0 0 5cm 0},clip,width=0.23\linewidth]{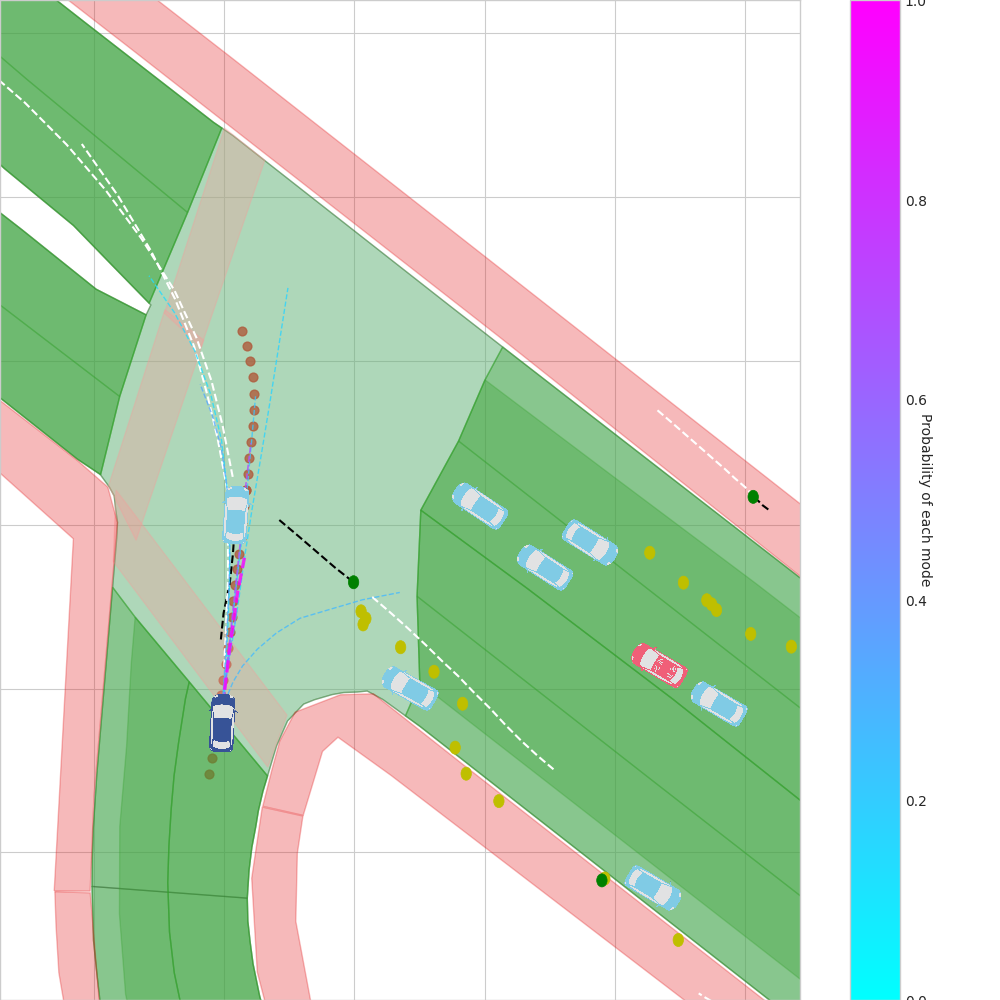} }
         {\includegraphics[trim={0 0 5cm 0},clip,width=0.23\linewidth]{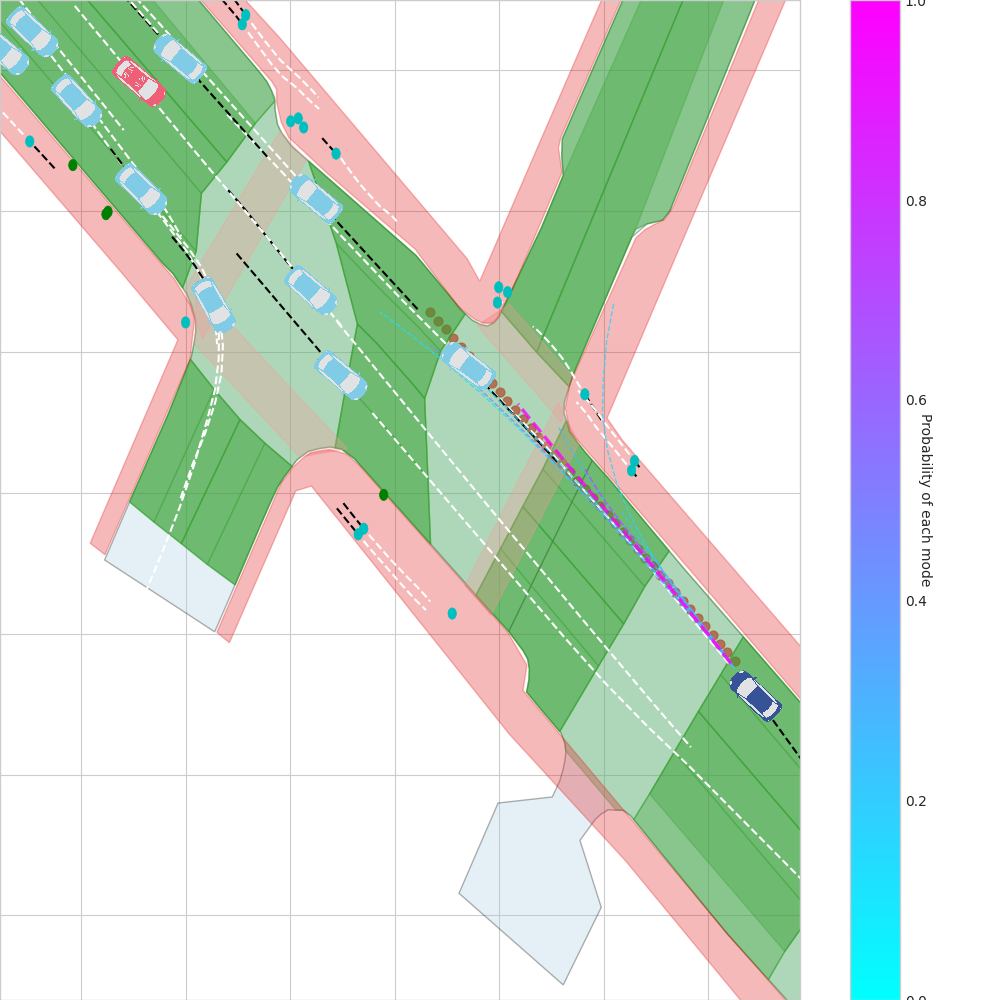}}
         {\includegraphics[trim={0 0 5cm 0},clip,width=0.23\linewidth]{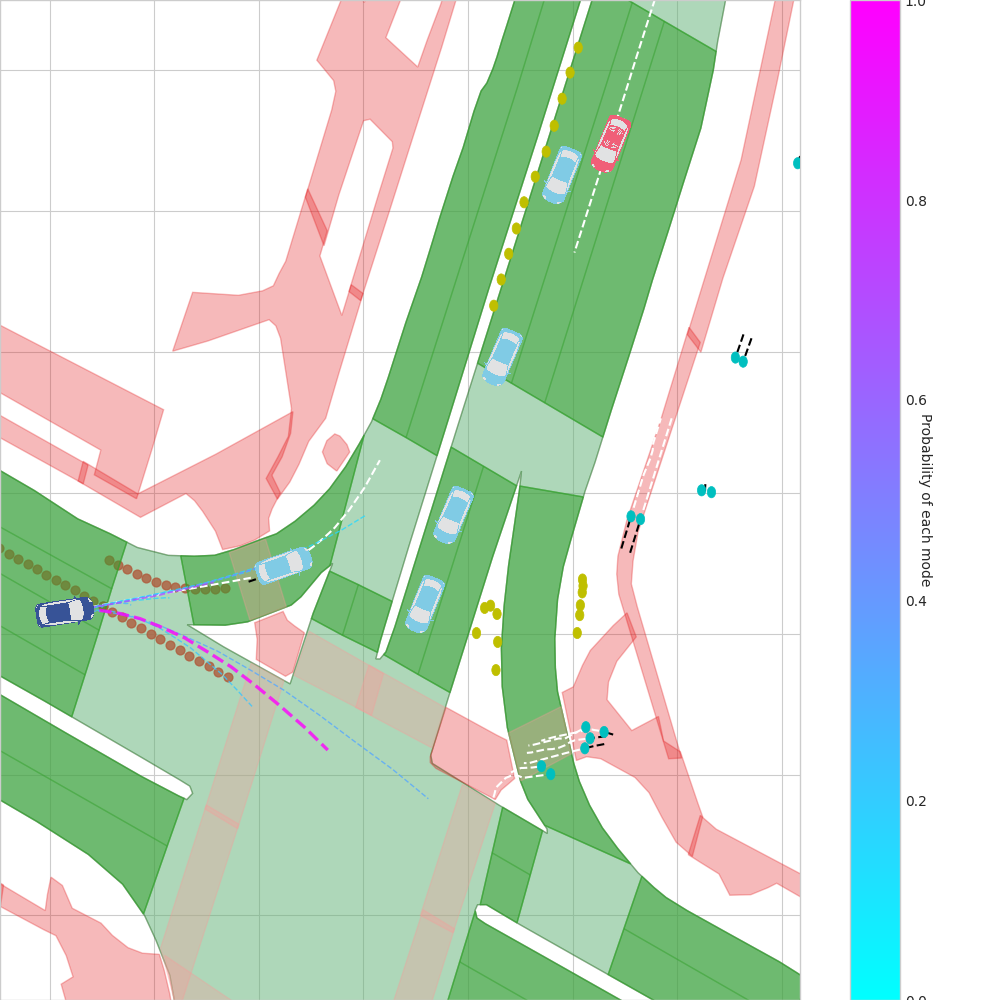}}
    \end{tabular}
    \caption{Qualitative assessment of lane ablation. Top row shows the predictions for 100\% recall. Bottom row shows the same scene where some critical lanes have been masked. Masked lanes are represented with faded red circles. In the first scene, most predictions take the alternative path. In the second scenario, the predictions follow the trajectory of the vehicle in front since the system cannot rely on lane information. In the third scenario, as there is no alternative route, predictions stay robust. In the last scenario, the most likely predictions turn right.  } 
    \label{fig:ablation2}
    
\end{figure*}

Figure \ref{fig:ablation2} shows four sample scenes. Masked lanes are indicated with faded red circles.
In the first scenario, the  probability of the alternative path -- continue straight -- increases, with most modes going in that direction. Only one mode with low probability turns right.
In the second scenario, when the whole ground-truth path is masked, the speed in most predictions is drastically decreased.  It seems that the prediction relies on the trajectory of the vehicle in front of the focal agent, since it does not take into account the information of the road topology.
In the third scenario, there is no other plausible path apart from the ground-truth. Although the predictions change slightly, they still comply to the traffic rules. The least likely goal goes off the road as it is unable to make a safe right turn. This is because the model can only be based on the right lane, which is separated from the current lane.  
Finally, in the last scenario, when the two lanes of the two plausible paths are masked, the most likely predictions are to turn right at the intersection rather than enter the merging lane, seeking the only alternative route where the lane is available. 
The results suggest that the model is robust to failures in the detection of some lane segments, providing diverse and sensible predictions that comply with the rules of the road.

\section{Discussion and Conclusions}
\label{sec:conclusions}

In this work, we advance toward explainable motion prediction by means of different explainability approaches. First, we propose a new approach -- Explainable Heterogeneous Graph-based Policy (XHGP) -- based on lane-graph traversals~\cite{deo2021multimodal} where we model the traffic scene as an heterogeneous graph and embeddings are learned using object-level and type-level attention mechanisms among the different interactions in the scene. This attention provides us with information about the most important agents and interactions in the scene for the final prediction. Next, we explore the same idea using GNNExplainer~\cite{GNNEXp}. Finally, we perform counterfactual reasoning to explain the predictions of individual instances. 
This analysis of different explainability techniques is a first step towards more transparent and trustworthy human-vehicle interactions, which is important not only from the point of view of the user, but also for developers debugging the system, and for certification and auditing procedures. 
With the increasing use of GNNs in different fields, more and more explainability techniques are appearing in the literature. The next steps aim to use these new techniques to better explain GNN-based motion prediction systems.

\section*{Acknowledgment}
This work was supported by HUMAINT project at the Digital Economy Unit at the Directorate-General Joint Research Centre (JRC) of the European Commission, and in part by the Spanish Ministry of Science and Innovation under Grant PID2020-114924RB-I00, and by the Community Region of Madrid under Grant S2018/EMT-4362 SEGVAUTO 4.0-CM. In addition, part of this work has been carried out during the \textit{Eye for AI Program} thanks to the support of Zenseact and AI Sweden.

\section*{Disclaimer}
The information and views expressed in this paper are purely those of the authors and do not necessarily reflect an official position of the European Commission.

\ifCLASSOPTIONcaptionsoff
  \newpage
\fi

\bibliographystyle{unsrt}

\begin{thebibliography}{10}

\bibitem{Bahari2021}
Mohammadhossein Bahari, Ismail Nejjar, and Alexandre Alahi.
\newblock Injecting knowledge in data-driven vehicle trajectory predictors.
\newblock {\em Transportation Research Part C: Emerging Technologies},
  128:103010, 2021.

\bibitem{Ghorai2022}
Prasenjit Ghorai, Azim Eskandarian, Young-Keun Kim, and Goodarz Mehr.
\newblock State estimation and motion prediction of vehicles and vulnerable
  road users for cooperative autonomous driving: A survey.
\newblock {\em IEEE Transactions on Intelligent Transportation Systems},
  23(10):16983--17002, 2022.

\bibitem{bib:Carrasco2021}
S.~Carrasco, D.~Fernández Llorca, and M.~A. Sotelo.
\newblock Scout: Socially-consistent and understandable graph attention network
  for trajectory prediction of vehicles and vrus.
\newblock In {\em 2021 IEEE Intelligent Vehicles Symposium (IV)}, pages
  1501--1508, 2021.

\bibitem{deo2021multimodal}
Nachiket Deo, Eric Wolff, and Oscar Beijbom.
\newblock Multimodal trajectory prediction conditioned on lane-graph
  traversals.
\newblock In {\em 5th Annual Conference on Robot Learning}, 2021.

\bibitem{Gu2021DenseTNTET}
Junru Gu, Chen Sun, and Hang Zhao.
\newblock Densetnt: End-to-end trajectory prediction from dense goal sets.
\newblock {\em 2021 IEEE/CVF International Conference on Computer Vision
  (ICCV)}, pages 15283--15292, 2021.

\bibitem{GNN}
Thomas~N. Kipf and Max Welling.
\newblock Semi-supervised classification with graph convolutional networks.
\newblock In {\em International Conference on Learning Representations (ICLR)},
  2017.

\bibitem{GAT}
Petar Veličković, Guillem Cucurull, Arantxa Casanova, Adriana Romero, Pietro
  Liò, and Yoshua Bengio.
\newblock Graph attention networks.
\newblock In {\em International Conference on Learning Representations (ICLR)},
  2018.

\bibitem{Carrasco2022}
Sandra Carrasco~Limeros, Sylwia Majchrowska, Joakim Johnander, Christoffer
  Petersson, Miguel~Ángel Sotelo, and David~Fernández Llorca.
\newblock Towards trustworthy multi-modal motion prediction: Evaluation and
  interpretability.
\newblock {\em ArXiv}, abs/2210.16144, 2022.

\bibitem{AIAct}
EC.
\newblock Regulation of the european parliament and the council laying down
  harmonised rules on artificial intelligence (artificial intelligence act) and
  amending certain union legislative acts., 2021.
\newblock COM(2021) 206 final,
  \url{https://eur-lex.europa.eu/legal-content/EN/TXT/?uri=CELEX\%3A52021PC0206}.

\bibitem{US-AAA2022}
Office of~U.S. Senator Ron~Wyden.
\newblock Algorithmic accountability act of 2022, 2022.
\newblock 117th Congress 2D Session,
  \url{https://www.congress.gov/bill/117th-congress/house-bill/6580/text}.

\bibitem{HLEGAI}
European Commission, Content Directorate-General~for Communications~Networks,
  and Technology.
\newblock {\em Ethics guidelines for trustworthy AI}.
\newblock Publications Office, 2019.

\bibitem{Llorca2023}
D.~Fernández~Llorca and E.~Gómez.
\newblock Trustworthy artificial intelligence requirements in the autonomous
  driving domain.
\newblock {\em {COMPUTER}}, February 2023.

\bibitem{Izquierdo2022}
Rubén~Izquierdo Gonzalo, Carlota~Salinas Maldonado, Javier~Alonso Ruiz,
  Ignacio~Parra Alonso, David~Fernández Llorca, and Miguel~Ángel Sotelo.
\newblock Testing predictive automated driving systems: Lessons learned and
  future recommendations.
\newblock {\em IEEE Intelligent Transportation Systems Magazine}, 14(6):77--93,
  2022.

\bibitem{Llorca2021}
D.~Fernández~Llorca and E.~Gómez~Gutierrez.
\newblock Trustworthy autonomous vehicles.
\newblock {\em EUR 30942 EN, Publications Office of the European Union,
  Luxembourg}, JRC127051, 2021.

\bibitem{Huang2022}
Yanjun Huang, Jiatong Du, Ziru Yang, Zewei Zhou, Lin Zhang, and Hong Chen.
\newblock A survey on trajectory-prediction methods for autonomous driving.
\newblock {\em IEEE Transactions on Intelligent Vehicles}, 7(3):652--674, 2022.

\bibitem{gunning2016explainable}
D~Gunning.
\newblock Explainable artificial intelligence (xai) darpa-baa-16-53.
\newblock {\em Defense Advanced Research Projects Agency}, 2016.

\bibitem{towards-rigorous}
Finale Doshi-Velez and Been Kim.
\newblock Towards a rigorous science of interpretable machine learning, 2017.

\bibitem{article-rudin}
Cynthia Rudin.
\newblock Stop explaining black box machine learning models for high stakes
  decisions and use interpretable models instead.
\newblock {\em Nature Machine Intelligence}, 1:206--215, 05 2019.

\bibitem{explaining}
Leilani~H. Gilpin, David Bau, Ben~Z. Yuan, Ayesha Bajwa, Michael Specter, and
  Lalana Kagal.
\newblock Explaining explanations: An overview of interpretability of machine
  learning.
\newblock In {\em 2018 IEEE 5th International Conference on Data Science and
  Advanced Analytics (DSAA)}, pages 80--89, 2018.

\bibitem{explain}
Finale Doshi-Velez, Mason Kortz, Ryan Budish, Chris Bavitz, Sam Gershman, David
  O'Brien, Kate Scott, Stuart Schieber, James Waldo, David Weinberger, Adrian
  Weller, and Alexandra Wood.
\newblock Accountability of ai under the law: The role of explanation.
\newblock {\em ArXiv}, abs/1711.01134, 2017.

\bibitem{GRIP2019}
Xin Li, Xiaowen Ying, and Mooi~Choo Chuah.
\newblock Grip: Graph-based interaction-aware trajectory prediction.
\newblock In {\em 2019 IEEE Intelligent Transportation Systems Conference
  (ITSC)}, pages 3960--3966, 2019.

\bibitem{Zhou2021}
Yutao Zhou, Huayi Wu, Hongquan Cheng, Kunlun Qi, Kai Hu, Chaogui Kang, and Jie
  Zheng.
\newblock Social graph convolutional lstm for pedestrian trajectory prediction.
\newblock {\em IET Intelligent Transport Systems}, 15(3):396--405, 2021.

\bibitem{Mo2021}
Xiaoyu Mo, Yang Xing, and Chen Lv.
\newblock Graph and recurrent neural network-based vehicle trajectory
  prediction for highway driving.
\newblock In {\em 2021 IEEE International Intelligent Transportation Systems
  Conference (ITSC)}, pages 1934--1939, 2021.

\bibitem{Tang2022}
Luqi Tang, Fuwu Yan, Bin Zou, Wenbo Li, Chen Lv, and Kewei Wang.
\newblock Trajectory prediction for autonomous driving based on multiscale
  spatial-temporal graph.
\newblock {\em IET Intelligent Transport Systems}, 2022.

\bibitem{Zhang2022-T-IV}
Kunpeng Zhang, Liang Zhao, Chengxiang Dong, Lan Wu, and Liang Zheng.
\newblock {AI-TP: Attention-based Interaction-aware Trajectory Prediction for
  Autonomous Driving}.
\newblock {\em IEEE Transactions on Intelligent Vehicles}, 2022.

\bibitem{Zhang2022-T-ITS}
Kunpeng Zhang, Xiaoliang Feng, Lan Wu, and Zhengbing He.
\newblock Trajectory prediction for autonomous driving using spatial-temporal
  graph attention transformer.
\newblock {\em IEEE Transactions on Intelligent Transportation Systems},
  23(11):22343--22353, 2022.

\bibitem{SpAGNN}
Sergio Casas, Cole Gulino, Renjie Liao, and Raquel Urtasun.
\newblock Spagnn: Spatially-aware graph neural networks for relational behavior
  forecasting from sensor data.
\newblock In {\em IEEE International Conference on Robotics and Automation
  (ICRA)}, pages 9491--9497, 05 2020.

\bibitem{vectornet2020}
Jiyang Gao, Chen Sun, Hang Zhao, Yi~Shen, Dragomir Anguelov, Congcong Li, and
  Cordelia Schmid.
\newblock Vectornet: Encoding hd maps and agent dynamics from vectorized
  representation.
\newblock In {\em 2020 IEEE/CVF Conference on Computer Vision and Pattern
  Recognition (CVPR)}, pages 11522--11530, 2020.

\bibitem{tnt}
Hang Zhao, Jiyang Gao, Tian Lan, Chen Sun, Benjamin Sapp, Balakrishnan
  Varadarajan, Yue Shen, Yi~Shen, Yuning Chai, Cordelia Schmid, Congcong Li,
  and Dragomir Anguelov.
\newblock Tnt: Target-driven trajectory prediction.
\newblock {\em ArXiv}, abs/2008.08294, 2020.

\bibitem{Zhao2022}
Cong Zhao, Andi Song, Yuchuan Du, and Biao Yang.
\newblock Trajgat: A map-embedded graph attention network for real-time vehicle
  trajectory imputation of roadside perception.
\newblock {\em Transportation Research Part C: Emerging Technologies},
  142:103787, 2022.

\bibitem{Khandelwal2020WhatIfMP}
Siddhesh Khandelwal, William Qi, Jagjeet Singh, Andrew Hartnett, and Deva
  Ramanan.
\newblock What-if motion prediction for autonomous driving.
\newblock {\em ArXiv}, abs/2008.10587, 2020.

\bibitem{Lanegcn}
Ming Liang, Bin Yang, Rui Hu, Yun Chen, Renjie Liao, Song Feng, and Raquel
  Urtasun.
\newblock Learning lane graph representations for motion forecasting.
\newblock In {\em European Conference on Computer Vision (ECCV)}, 2020.

\bibitem{LaneRCNN}
Wenyuan Zeng, Ming Liang, Renjie Liao, and Raquel Urtasun.
\newblock Lanercnn: Distributed representations for graph-centric motion
  forecasting.
\newblock In {\em IEEE/RSJ International Conference on Intelligent Robots and
  Systems (IROS)}, pages 532--539, 09 2021.

\bibitem{Yuan_2020}
Hao Yuan, Jiliang Tang, Xia Hu, and Shuiwang Ji.
\newblock {XGNN}: Towards model-level explanations of graph neural networks.
\newblock In {\em Proceedings of the 26th {ACM} {SIGKDD} International
  Conference on Knowledge Discovery and Data Mining}. {ACM}, aug 2020.

\bibitem{8954227}
Phillip~E. Pope, Soheil Kolouri, Mohammad Rostami, Charles~E. Martin, and Heiko
  Hoffmann.
\newblock Explainability methods for graph convolutional neural networks.
\newblock In {\em 2019 IEEE/CVF Conference on Computer Vision and Pattern
  Recognition (CVPR)}, pages 10764--10773, 2019.

\bibitem{SA}
Federico Baldassarre and Hossein Azizpour.
\newblock Explainability techniques for graph convolutional networks.
\newblock {\em ArXiv}, abs/1905.13686, 2019.

\bibitem{GNNEXp}
Rex Ying, Dylan Bourgeois, Jiaxuan You, Marinka Zitnik, and Jure Leskovec.
\newblock {GNNExplainer: Generating Explanations for Graph Neural Networks}.
\newblock In {\em 33rd Conference on Neural Information Processing Systems
  (NeurIPS)}, 2019.

\bibitem{PGExplanier}
Dongsheng Luo, Wei Cheng, Dongkuan Xu, Wenchao Yu, Bo~Zong, Haifeng Chen, and
  Xiang Zhang.
\newblock Parameterized explainer for graph neural network.
\newblock In {\em 34th International Conference on Neural Information
  Processing Systems (NeurIPS)}, 2020.

\bibitem{GraphMask}
Michael~Sejr Schlichtkrull, Nicola~De Cao, and Ivan Titov.
\newblock Interpreting graph neural networks for {\{}nlp{\}} with
  differentiable edge masking.
\newblock In {\em International Conference on Learning Representations (ICLR)},
  2021.

\bibitem{casual}
Wanyu Lin, Hao Lan, and Baochun Li.
\newblock Generative causal explanations for graph neural networks.
\newblock In {\em 38th International Conference on Machine Learning}, 2021.

\bibitem{CFGNN}
Ana Lucic, Maartje ter Hoeve, Gabriele Tolomei, Maarten de~Rijke, and Fabrizio
  Silvestri.
\newblock {CF-GNNExplainer: Counterfactual Explanations for Graph Neural
  Networks}.
\newblock In {\em 25th International Conference on Artificial Intelligence and
  Statistics}, 2021.

\bibitem{Schnake_2021}
Thomas Schnake, Oliver Eberle, Jonas Lederer, Shinichi Nakajima, Kristof~T.
  Schütt, Klaus-Robert Müller, and Grégoire Montavon.
\newblock Higher-order explanations of graph neural networks via relevant
  walks.
\newblock {\em IEEE Transactions on Pattern Analysis and Machine Intelligence},
  44(11):7581--7596, 2022.

\bibitem{decomposition}
Robert Schwarzenberg, Marc H{\"u}bner, David Harbecke, Christoph Alt, and
  Leonhard Hennig.
\newblock Layerwise relevance visualization in convolutional text graph
  classifiers.
\newblock In {\em 13th Workshop on Graph-Based Methods for Natural Language
  Processing (TextGraphs-13)}, pages 58--62, 2019.

\bibitem{PGMExp}
Minh~N. Vu and My~T. Thai.
\newblock {PGM-Explainer: Probabilistic Graphical Model Explanations for Graph
  Neural Networks}.
\newblock In {\em 34th Conference on Neural Information Processing Systems
  (NeurIPS)}, 2020.

\bibitem{GraphLime}
Qiang Huang, Makoto Yamada, Yuan Tian, Dinesh Singh, Dawei Yin, and Yi~Chang.
\newblock {GraphLIME: Local Interpretable Model Explanations for Graph Neural
  Networks}.
\newblock {\em ArXiv}, abs/2001.06216, 2020.

\bibitem{ig}
Mukund Sundararajan, Ankur Taly, and Qiqi Yan.
\newblock Axiomatic attribution for deep networks.
\newblock In {\em Proceedings of the 34th International Conference on Machine
  Learning - Volume 70}, ICML'17, page 3319–3328. JMLR.org, 2017.

\bibitem{Zhang2022-TRC}
Kunpeng Zhang and Li~Li.
\newblock Explainable multimodal trajectory prediction using attention models.
\newblock {\em Transportation Research Part C: Emerging Technologies},
  143:103829, 2022.

\bibitem{Zhang2022}
Qingzhao Zhang, Shengtuo Hu, Jiachen Sun, Qi~Alfred Chen, and Z.~Morley Mao.
\newblock On adversarial robustness of trajectory prediction for autonomous
  vehicles.
\newblock In {\em 2022 IEEE/CVF Conference on Computer Vision and Pattern
  Recognition (CVPR)}, pages 15138--15147, 2022.

\bibitem{iehgcn}
Yaming Yang, Ziyu Guan, Jianxin Li, Wei Zhao, Jiangtao Cui, and Quan Wang.
\newblock Interpretable and efficient heterogeneous graph convolutional
  network.
\newblock {\em IEEE Transactions on Knowledge and Data Engineering}, pages
  1--1, 2021.

\bibitem{DBLP:conf/iclr/KipfW17}
Thomas~N. Kipf and Max Welling.
\newblock Semi-supervised classification with graph convolutional networks.
\newblock In {\em 5th International Conference on Learning Representations,
  {ICLR} 2017, Toulon, France, April 24-26, 2017, Conference Track
  Proceedings}. OpenReview.net, 2017.

\bibitem{NIPS2017_3f5ee243}
Ashish Vaswani, Noam Shazeer, Niki Parmar, Jakob Uszkoreit, Llion Jones,
  Aidan~N Gomez, \L~ukasz Kaiser, and Illia Polosukhin.
\newblock Attention is all you need.
\newblock In I.~Guyon, U.~Von Luxburg, S.~Bengio, H.~Wallach, R.~Fergus,
  S.~Vishwanathan, and R.~Garnett, editors, {\em Advances in Neural Information
  Processing Systems}, volume~30. Curran Associates, Inc., 2017.

\bibitem{Jaume2020TowardsEG}
Guillaume Jaume, Pushpak Pati, Antonio Foncubierta-Rodr{\'i}guez, Florinda
  Feroce, G.~Scognamiglio, Anna~Maria Anniciello, Jean-Philippe Thiran, Orcun
  Goksel, and Maria Gabrani.
\newblock Towards explainable graph representations in digital pathology.
\newblock {\em ArXiv}, abs/2007.00311, 2020.

\bibitem{Rao2021QuantitativeEO}
Jiahua Rao, Shuangjia Zheng, and Yuedong Yang.
\newblock Quantitative evaluation of explainable graph neural networks for
  molecular property prediction.
\newblock {\em ArXiv}, abs/2107.04119, 2021.

\bibitem{patology}
Guillaume Jaume, Pushpak Pati, Behzad Bozorgtabar, Antonio Foncubierta,
  Anna~Maria Anniciello, Florinda Feroce, Tilman Rau, Jean-Philippe Thiran,
  Maria Gabrani, and Orcun Goksel.
\newblock Quantifying explainers of graph neural networks in computational
  pathology.
\newblock In {\em 2021 IEEE/CVF Conference on Computer Vision and Pattern
  Recognition (CVPR)}, pages 8102--8112, 2021.

\bibitem{Zhdanov2022InvestigatingBC}
Maksim Zhdanov, Saskia Steinmann, and Nico Hoffmann.
\newblock Investigating brain connectivity with graph neural networks and
  gnnexplainer.
\newblock {\em ArXiv}, abs/2206.01930, 2022.

\bibitem{nuscenes2019}
Holger Caesar, Varun Bankiti, Alex~H. Lang, Sourabh Vora, Venice~Erin Liong,
  Qiang Xu, Anush Krishnan, Yu~Pan, Giancarlo Baldan, and Oscar Beijbom.
\newblock nuscenes: A multimodal dataset for autonomous driving.
\newblock {\em arXiv preprint arXiv:1903.11027}, 2019.

\bibitem{Yuan2022ExplainabilityIG}
Hao Yuan, Haiyang Yu, Shurui Gui, and Shuiwang Ji.
\newblock Explainability in graph neural networks: A taxonomic survey.
\newblock {\em IEEE Transactions on Pattern Analysis and Machine Intelligence},
  pages 1--19, 2022.

\bibitem{traj++}
Tim Salzmann, Boris Ivanovic, Punarjay Chakravarty, and Marco Pavone.
\newblock Trajectron++: Dynamically-feasible trajectory forecasting with
  heterogeneous data.
\newblock In {\em European Conference on Computer Vision (ECCV)}, 2020.

\bibitem{cxx}
Chenxu Luo, Lin Sun, Dariush Dabiri, and Alan Yuille.
\newblock Probabilistic multi-modal trajectory prediction with lane attention
  for autonomous vehicles.
\newblock In {\em IEEE/RSJ International Conference on Intelligent Robots and
  Systems (IROS)}, pages 2370--2376, 10 2020.

\bibitem{ptp}
Nachiket Deo and Mohan Trivedi.
\newblock Trajectory forecasts in unknown environments conditioned on
  grid-based plans.
\newblock {\em ArXiv}, abs/2001.00735, 01 2020.

\bibitem{gilles2022thomas}
Thomas Gilles, Stefano Sabatini, Dzmitry Tsishkou, Bogdan Stanciulescu, and
  Fabien Moutarde.
\newblock {THOMAS}: Trajectory heatmap output with learned multi-agent
  sampling.
\newblock In {\em International Conference on Learning Representations}, 2022.

\end{thebibliography}

\end{document}